%% file: main.tex
\documentclass[10pt,twocolumn,letterpaper]{article}

\usepackage[accsupp]{axessibility}  % Improves PDF readability for those with disabilities.

\usepackage[pagenumbers]{cvpr} % To force page numbers, e.g. for an arXiv version


\definecolor{cvprblue}{rgb}{0.21,0.49,0.74}
\usepackage[pagebackref,breaklinks,colorlinks,allcolors=cvprblue]{hyperref}

\usepackage{url}

\usepackage{graphicx}
\usepackage{caption}
\usepackage{subcaption}
\usepackage{afterpage}
\usepackage{placeins}
\usepackage{amsmath}
\usepackage{lipsum} % for dummy text
\usepackage{tabularx}
\usepackage{multirow}
\usepackage{enumitem}
\usepackage[ruled,vlined]{algorithm2e}
\usepackage{booktabs}
\usepackage{makecell}

\def\paperID{10865} % *** Enter the Paper ID here
\def\confName{CVPR}
\def\confYear{2025}

\title{Towards Better Alignment: Training Diffusion Models with Reinforcement Learning Against Sparse Rewards}

\author{
Zijing Hu$^{1}$\thanks{Equal contribution. $^{\dag}$Corresponding author.}
\quad
Fengda Zhang$^{2\,*}$
\quad
Long Chen$^{3}$
\quad
Kun Kuang$^{1\,\dag}$
\quad
Jiahui Li$^{1}$
\quad
Kaifeng Gao$^{1}$
\\
Jun Xiao$^{1}$
\quad
Xin Wang$^{4}$
\quad
Wenwu Zhu$^{4}$
\\ {\small
$^1$Zhejiang University,\!
$^2$Nanyang Technological University,\!
$^3$The Hong Kong University of Science and Technology,\!
$^4$Tsinghua University
}
\\ {\tt\small
\{zj.hu,fdzhang\}@zju.edu.cn, 
zjuchenlong@gmail.com, 
\{kunkuang,jiahuil,kite\_phone\}@zju.edu.cn,}  \\{\tt\small
junx@cs.zju.edu.cn, 
\{xin\_wang,wwzhu\}@tsinghua.edu.cn
}
}

\begin{document}

\maketitle

\input{Section/abstract}
\input{Section/introduction}
\input{Section/relatedwork}
\input{Section/method}
\input{Section/experiment}
\input{Section/conclusion}

{
    \small
    \bibliographystyle{ieeenat_fullname}
    \bibliography{main}
}

\clearpage
\appendix
\input{Section/appendix}

\end{document}

%% file: Section/abstract.tex
\begin{abstract}

Diffusion models have achieved remarkable success in text-to-image generation. However, their practical applications are hindered by the misalignment between generated images and corresponding text prompts. To tackle this issue, reinforcement learning (RL) has been considered for diffusion model fine-tuning. Yet, RL's effectiveness is limited by the challenge of sparse reward, where feedback is only available at the end of the generation process. This makes it difficult to identify which actions during the denoising process contribute positively to the final generated image, potentially leading to ineffective or unnecessary denoising policies. 
To this end, this paper presents a novel RL-based framework that addresses the sparse reward problem when training diffusion models.
Our framework, named $\text{B}^2\text{-DiffuRL}$, employs two strategies: \textbf{B}ackward progressive training and \textbf{B}ranch-based sampling.
For one thing, backward progressive training focuses initially on the final timesteps of denoising process and gradually extends the training interval to earlier timesteps, easing the learning difficulty from sparse rewards. For another, we perform branch-based sampling for each training interval. By comparing the samples within the same branch, we can identify how much the policies of the current training interval contribute to the final image, which helps to learn effective policies instead of unnecessary ones.
$\text{B}^2\text{-DiffuRL}$ is compatible with existing optimization algorithms. 
Extensive experiments demonstrate the effectiveness of $\text{B}^2\text{-DiffuRL}$ in improving prompt-image alignment and maintaining diversity in generated images.
The code for this work is available\footnote{\url{https://github.com/hu-zijing/B2-DiffuRL}.}. 

\end{abstract}

%% file: Section/introduction.tex
\begin{figure}
    \vspace{-0.6em}
    \centering
    \setlength{\abovecaptionskip}{0.2em}
    \includegraphics[width=1.0\linewidth]{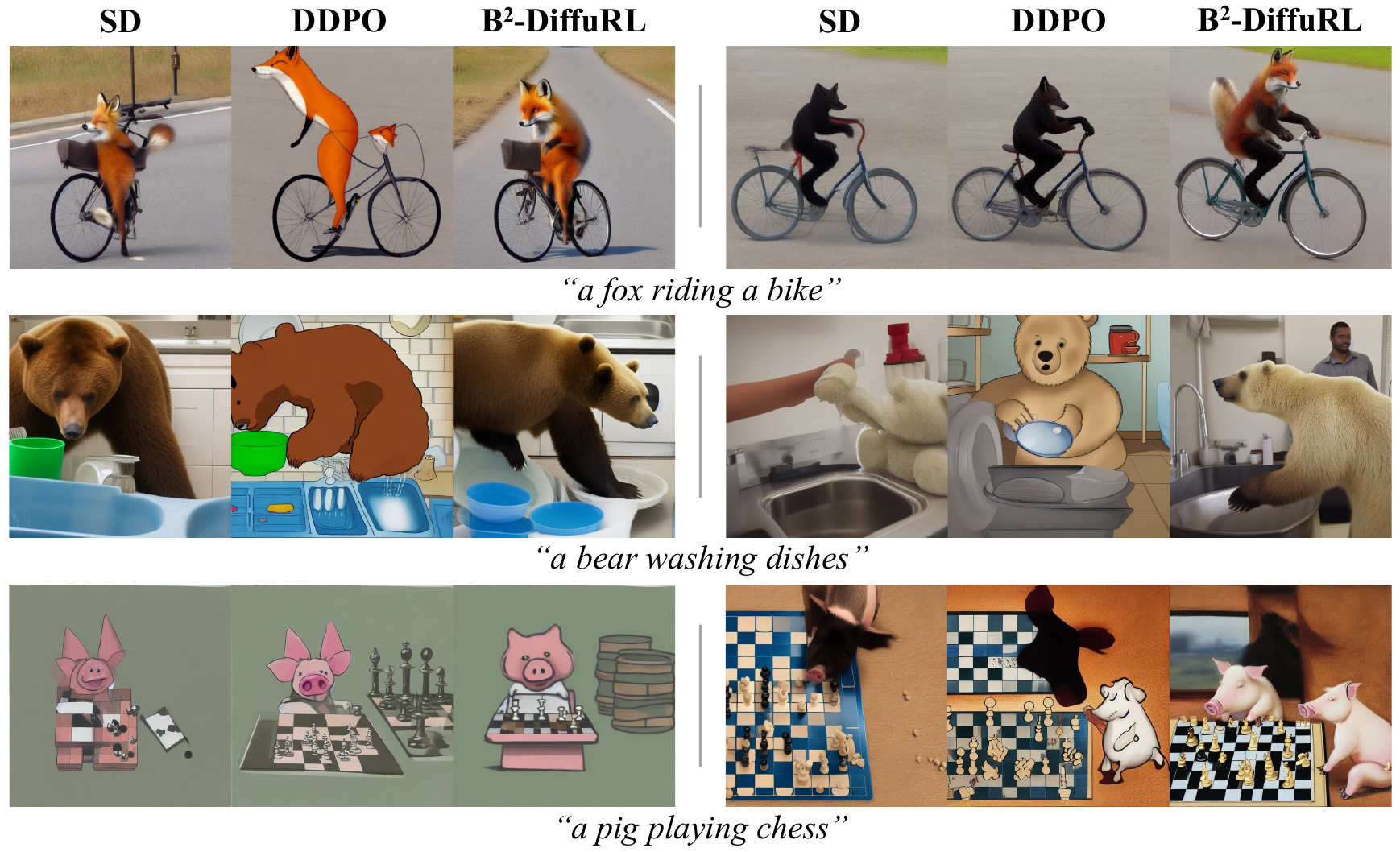}
    \caption{\textbf{(Prompt-image Misalignment)} Text-to-image diffusion models (\textit{e.g.}, Stable Diffusion (SD)~\cite{rombach2022latent}) may not generate high-quality images that accurately align with prompts. Existing reinforcement learning-based diffusion model fine-tuning methods (\textit{e.g.}, DDPO~\cite{Black2023TrainingDM}) have limited effect and loss of image diversity. For each set of images above, we use the same seed for sampling.}
    \label{fig1}
    \vspace{-1.2em}
\end{figure}

\vspace{-1em}
\section{Introduction}

\begin{figure*}
  \centering
  \setlength{\abovecaptionskip}{0.2em}
  \includegraphics[width=1.0\textwidth]{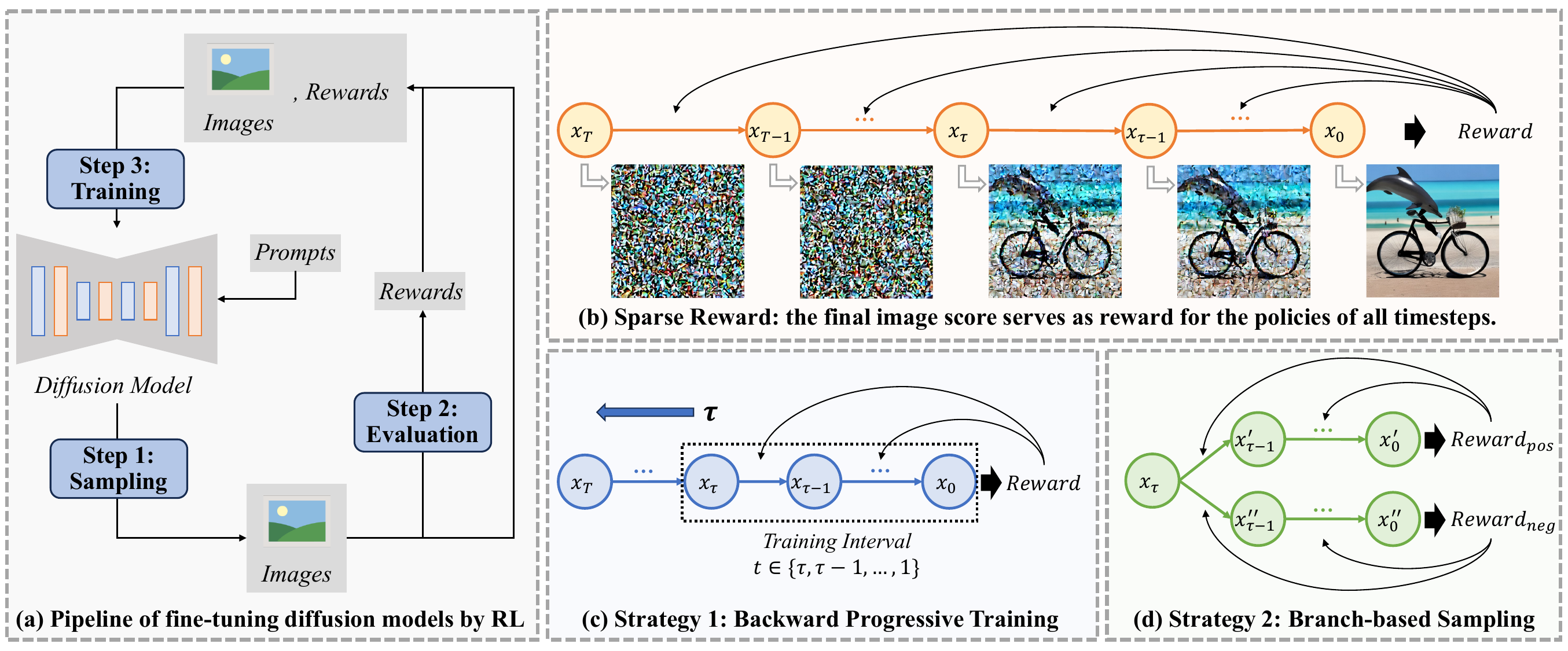}
  \caption{(\textbf{Sparse Reward}) When people train diffusion models with reinforcement learning (RL), the reward is only available at the end of the generation process. This sparsity limits the success of RL in diffusion models. We propose $\text{B}^2\text{-DiffuRL}$, a new RL framework with two strategies, to mitigate this issue.} 
  \label{fig2}
  \vspace{-1.4em}
\end{figure*}

The text-to-image generation task aims to produce images from textual descriptions, holding significant potential for various applications~\cite{ramesh2021dalle,saharia2022photorealistic}. Recently, diffusion models have garnered widespread attention due to their success in this domain~\cite{ho2020denoising,song2020improved,dhariwal2021diffusion}. These models employ a sequential denoising process that transforms random noise into detailed images. However, even the most advanced text-to-image diffusion models, such as DALLE3~\cite{betker2023improving} and Stable Diffusion~\cite{rombach2022latent}, often encounter issues with misalignment between the generated images and the textual descriptions~\cite{jiang2024comat}. This misalignment limits the practicality and effectiveness of these models in real-world applications.

To solve this problem, recent studies have explored incorporating reinforcement learning (RL) techniques to fine-tune pre-trained text-to-image diffusion models~\cite{lee2023aligning,xu2023imagereward,fan2023dpok,wallace2023diffusion,Black2023TrainingDM,prabhudesai2023aligning}. By formulating the step-by-step denoising process as a \textit{sequential decision-making problem}, RL enables diffusion models to optimize for specific long-term objectives, beyond merely fitting to static data as done in standard supervised learning~\cite{kaelbling1996reinforcement,wang2024patch,rombach2022latent}. In this formulation, noisy images at different timesteps are viewed as \textit{states} in RL, while denoising at each timestep corresponds to an \textit{action}. The alignment scores between the final generated images and the textual prompts, which can be derived from human preferences or model evaluations, serve as the \textit{rewards}. The pipeline of training diffusion models with RL is illustrated in Figure~\ref{fig2}~(a). Researchers first sample images using the diffusion model with given prompts and then calculate the alignment scores as rewards. These sampled trajectories, consisting of images at different timesteps and their corresponding alignment scores, can be used as training data for RL to further enhance the diffusion models~\cite{hoi2018online}.

However, RL has so far made limited success in improving prompt-image alignment, primarily due to the key challenge of \textbf{\textit{sparse reward}}. As shown in Figure~\ref{fig2}~(b), reward in this context is sparse because it is only available at the end of the generation process. Sparse rewards are harmful to RL-based diffusion fine-tuning in two ways:
\begin{itemize}
\item \textbf{\textit{Limited improvement in alignment}}. The denoising actions at different timesteps focus on varying levels of semantics (\textit{e.g.}, early timesteps define layout, middle timesteps refine style, and late timesteps enhance detailed objects) and have different impacts on the final image~\cite{yue2024few,yue2024exploring}.
With sparse rewards, it is difficult to identify which actions during the denoising process contribute positively to the final alignment, so actions at different timesteps receive inappropriate rewards. As a result, learning effective policies becomes challenging. 
\item \textbf{\textit{Sacrificing diversity for better alignment}}. To achieve higher alignment score, the model may learn unnecessary policies. For example, with prompts like ``\textit{a bear washing dishes}'', cartoon-like images are more likely to get higher rewards than realistic photographs because the prompts are often depicted in a cartoon style in pre-training data.
With sparse rewards, model fine-tuned via naive RL algorithms (\textit{e.g.}, DDPO~\cite{Black2023TrainingDM}) may learn these unnecessary policies about styles, resulting in generating only cartoon-like images, as shown in Figure~\ref{fig1}. This shows a trade-off between alignment and diversity, where alignment is improved at the expense of diversity~\cite{zhang2024inherent,sadat2023cads}.
\end{itemize}
The challenge of sparse reward has attracted widespread attention in traditional RL~\cite{trott2019keeping,hare2019dealing}. The classic solutions are constructing additional rewards by various techniques, such as reward shaping~\cite{riedmiller2018learning,memarian2021self}, to achieve dense reward functions~\cite{devidze2022exploration,ng1999policy,hu2020learning,gupta2022unpacking}. 
Unfortunately, these solutions are not suitable for diffusion models because it is hard to evaluate the noisy images in the denoising process. This motivates us to ask: \textit{How can we mitigate the negative effects of sparse rewards when using RL to train diffusion models}?

In this paper, we introduce a novel RL-based fine-tuning framework for diffusion-based text-to-image generation to address the challenge of sparse reward, which we refer to as \textbf{$\text{B}^2\text{-DiffuRL}$}~\footnote{$\text{B}^2\text{-DiffuRL}$ is short for \textbf{B}ackward progressive training and \textbf{B}ranch-based sampling for \textbf{R}einforcement \textbf{L}earning in \textbf{Diffu}sion models.}.
Our framework employs two strategies. The first one is \textbf{\textit{backward progressive training}} (BPT), applied to the training stage.
Initially, we focus training on only the final timesteps of the image generation process, as shown in Figure~\ref{fig2}~(c). As training rounds increase, we gradually extend the training interval backward to cover all timesteps, and achieve training on the entire denoising process in the end. 
The second strategy is \textbf{\textit{branch-based sampling}} (BS), applied to the sampling stage.
For each training interval in denosing process, we perform branch sampling to get multiple samples under each branch, as shown in Figure~\ref{fig2}~(d). Within each branch, we only select the best and worst samples to form a contrastive sample pair for RL training. 

Our framework has the following three capabilities: 
(1) \textbf{\textit{Better prompt-image alignment}}. 
With small training interval, BPT strategy enables the models to easily and quickly learn the policies for the later timesteps of generation. As the model becomes proficient in these later timesteps, it progressively learns to manage the earlier timesteps of the denoising process. By mitigating the complexity of dealing with the entire process from the outset, BPT reduces the learning difficulty associated with sparse rewards. 
Moreover, with BS strategy, the contrastive samples within the same branch share identical states and actions up to the start of the training interval.
By comparing the contrastive samples, the models can accurately identify how much the denosing policies of the current training interval contribute to the final image during training. 
(2) \textbf{\textit{Maintaining diversity when improving alignment}}. 
Denoised from the same intermediate state, the contrastive samples share similar coarse-grained visual information (\textit{e.g.}, image styles) but receive different rewards. It prevents the models from learning unnecessary policies (\textit{e.g.}, about image styles) as shortcuts to achieve high rewards, thus helping maintain diversity.
(3) \textbf{\textit{Compatibility}}. Although we mainly compare with the current state-of-the-art RL-based fine-tuning algorithm called DDPO~\cite{Black2023TrainingDM} in this paper, our framework is compatible with any previous optimization algorithm such as policy gradient~\cite{schulman2017ppo}, DPO~\cite{rafailov2023direct, wallace2023diffusion} and DPOK~\cite{fan2023dpok}. Experiments show that applying $\text{B}^2\text{-DiffuRL}$ can improve effectiveness of different algorithms in terms of both alignment and diversity.

Our contributions can be summarized as:
(1) We investigate the problem of RL-based diffusion models fine-tuning for improving prompt-image alignment, and for the first time highlight the challenge of sparse reward.
(2) We propose a compatible RL-based fine-tuning framework named $\text{B}^2\text{-DiffuRL}$, employing backward progressive training and branch-based sampling strategies, to address the above challenge.
(3) Extensive experimental results on Stable Diffusion~\cite{rombach2022latent} show the effectiveness of $\text{B}^2\text{-DiffuRL}$ in terms of both alignment and diversity when compatible with different RL algorithms, without increasing computational cost.

%% file: Section/relatedwork.tex
\section{Related Work}
\subsection{Text-to-Image Diffusion Models}
Diffusion models have gained substantial attention for their ability to generate high-quality samples~\cite{ho2020denoising,song2020improved,yang2023diffusion,sultan1990meta}. One of the primary applications of diffusion models is image generation~\cite{ho2022cascaded,batzolis2021conditional}. These models have been shown to produce images that are both high in fidelity and diversity, rivaling the outputs of Generative Adversarial Networks (GANs)~\cite{dhariwal2021diffusion,goodfellow2020generative}. The extension of diffusion models to text-to-image generation has opened up possibilities for creating images from textual descriptions~\cite{zhang2023text}. Works like DALL-E~\cite{ramesh2021dalle} and Imagen~\cite{saharia2022imagen} have demonstrated that diffusion models can be effectively conditioned on textual input to produce corresponding images. Despite their success, text-to-image diffusion models often suffer from the issue of prompt-image misalignment~\cite{kumari2023multi,nichol2021improved}. 

\subsection{Reinforcement Learning with Sparse Reward}
Reinforcement Learning (RL) is a learning paradigm in which an agent learns to make decisions by interacting with an environment to maximize cumulative rewards~\cite{kaelbling1996reinforcement,polydoros2017survey}.
Applications of RL span various domains, including gaming, robotics, finance, and healthcare~\cite{luong2019applications,coronato2020reinforcement}.
Recently, RL has played an important role in alignment. For example, RL has been leveraged to fine-tune large language models (LLMs), ensuring that the generated outputs align with human values and intentions~\cite{christiano2017deep}.
One of the significant challenges in RL is dealing with sparse rewards, where feedback signals are infrequent and the agent must explore extensively to discover rewarding states~\cite{riedmiller2018learning,trott2019keeping}. Traditional RL algorithms struggle in such settings due to the inefficiency in learning from limited feedback~\cite{memarian2021self,hare2019dealing}. Various techniques have been proposed to address this challenge~\cite{nachum2018data,andrychowicz2017hindsight}, such as reward shaping~\cite{devidze2022exploration,ng1999policy,hu2020learning,gupta2022unpacking}, where additional heuristic rewards are provided to guide the agent.
However, these classic RL strategies can not be applied to our problem directly, since it is difficult to evaluate the noisy images during denoising process.

\subsection{Improving Alignment of Diffusion Models}

Early diffusion models focused primarily on the quality and fidelity of the generated images~\cite{ho2020denoising,song2020improved,dhariwal2021diffusion}. 
However, as the demand for a more interactive and user-driven generation grew, improving alignment between prompts and generated images is crucial for enhancing the usability and reliability of these models in practical applications~\cite{zhang2023adding,ruiz2023dreambooth,li2022diffusion,epstein2023diffusion}.
The initial approaches to conditioning diffusion models on text prompts employ a variety of techniques, including both classifier guidance and classifier-free guidance~\cite{dhariwal2021diffusion, ho2021classifierfree}. 
With the advent of LDMs~\cite{rombach2022highresolution}, subsequent researches focus on fine-tuning pre-trained models to enhance alignment~\cite{hu2021lora,li2024aligning}. 
Recently, RL has been employed to fine-tune the text-to-image diffusion models~\cite{lee2023aligning,xu2023imagereward,fan2023dpok,wallace2023diffusion,Black2023TrainingDM,prabhudesai2023aligning,clark2024directlyfinetuningdiffusionmodels, Yang_2024_CVPR, yang2024adensereward}.
However, the issue of sparse rewards limits the performance of such methods in prompt-image alignment, and even sacrifices a lot of diversity in order to improve controllability. In this paper, by mitigating the negative effects of sparse rewards, we further develop the application of RL in training diffusion models.

%% file: Section/method.tex
\begin{figure*}
  \centering
  \setlength{\abovecaptionskip}{0.5em}
  \includegraphics[width=1.0\textwidth]{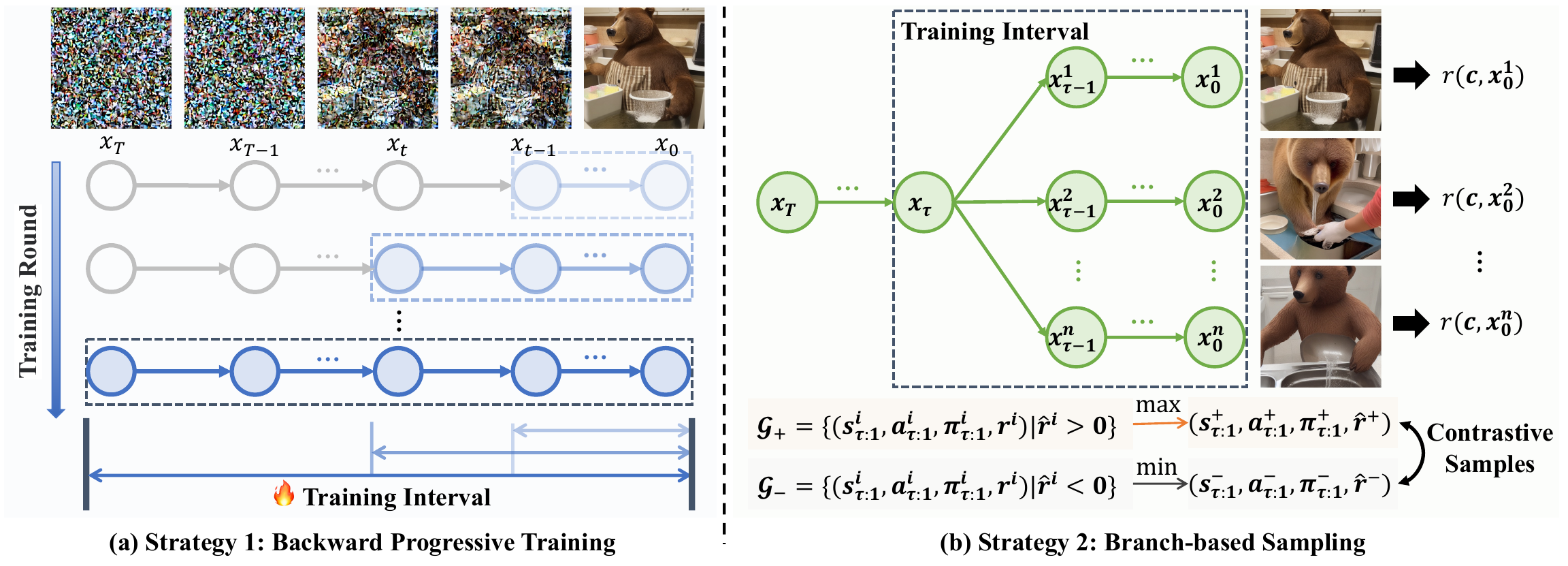}
  \caption{(\textbf{Method}) We propose the framework $\text{B}^2\text{-DiffuRL}$, employing two strategies to address the challenge of sparse rewards. (a) Backward progressive training strategy: We focus initially on the final timesteps of the denoising process and gradually extend the training interval to earlier timesteps, easing the learning difficulty associated with sparse rewards. (b) Branch-based sampling strategy: We perform branch-based sampling at the beginning of each training interval. Comparisons between samples within the same branch provide a clear indication of whether the policies of the current training interval positively contribute to the final images.} 
  \label{fig3}
\end{figure*}

\section{Method}

In this section, we first introduce how to train diffusion models with RL. Then we highlight the challenge of sparse reward in this context. Finally, we introduce $\text{B}^2\text{-DiffuRL}$, employing two strategies to address this challenge. 
$\text{B}^2\text{-DiffuRL}$ can be compatible with different RL algorithms, such as DDPO~\cite{Black2023TrainingDM}, DPO~\cite{wallace2023diffusion} and DPOK~\cite{fan2023dpok}. 

\subsection{Problem and Challenge}

\noindent\textbf{Text-to-Image Diffusion Models.}
Text-to-image diffusion models iteratively refine random noise into a coherent image that matches the given prompt~\cite{rombach2022latent}. The process of diffusion models consists of two phases: the forward process and the reverse process~\cite{ho2020denoising}.
In the forward process, an image $\mathbf{x}_0$ is gradually corrupted into pure noise $\mathbf{x}_T$ through $T$ steps, where Gaussian noise is added at each step.
The reverse process aims to generate an image from pure noise conditioned on a textual description $\mathbf{c}$ by denoising iteratively~\cite{ho2020denoising, song2022denoising}:
\begin{equation}
    p_\theta(\mathbf{x}_{t-1} \mid \mathbf{x}_t, \mathbf{c}) = \mathcal{N}(\mathbf{x}_{t-1}; \mu_\theta(\mathbf{x}_t, t, \mathbf{c}), \sigma_t \mathbf{I}^2),
\end{equation}
where $\mu_\theta$ is predicted by a diffusion model parameterized by $\theta$, and $\sigma_t$ is the fixed timestep-dependent variance.

\noindent\textbf{Training Diffusion Models with RL.}
The denoising process of diffusion models can be formulated as a \textit{sequential decision-making problem}. Therefore, this process can be viewed through the lens of RL, where each step in the denoising process is considered as a decision made by an agent (the diffusion model).
Following this formulation, the \textit{state} $s_t$ at each timestep is represented by $(\mathbf{c}, t, \mathbf{x}_t)$, \textit{i.e.}, the text prompt, the current timestep, and the noisy image at the current timestep. The sequence of states represents the gradual refinement from noise to the final image.
The \textit{action} $a_t$ at each timestep involves denoising by sampling the next noisy image $\mathbf{x}_{t-1}$.
The \textit{policy} $\pi_\theta$, parameterized by $\theta$, defines the action selection strategy. In this context, the policy is defined as $\pi_\theta(a_t \mid s_t) = p_\theta(\mathbf{x}_{t-1} \mid \mathbf{x}_t, \mathbf{c})$.
The \textit{reward} can be defined as a prompt-image alignment score $r(\mathbf{c}, \mathbf{x}_0) \in \mathbb{R}$, which is given by human preferences or model evaluations. A larger reward means a better prompt-image alignment.
To improve the prompt-image alignment of diffusion models, we can execute RL-based training by maximizing the following objective:
\begin{equation}
\mathcal{J}_{\text{RL}}(\theta) = \mathbb{E}_{\mathbf{c} \sim p(\mathbf{c}), \mathbf{x}_0 \sim p_{\theta}(\mathbf{x}_0 \mid \mathbf{c})} \left[ r(\mathbf{x}_0, \mathbf{c}) \right],
\label{loss}
\end{equation}
where $p(\mathbf{c})$ follows a uniform distribution, meaning that we randomly sample prompts from a candidate set of prompts.
To construct the training data for RL, we first collect denoising trajectories via sampling based on diffusion models.
Then we can update parameters $\theta$ via gradient descent~\cite{mohamed2020monte}.

\noindent\textbf{Challenge of Sparse Reward.}
However, the reward $r(\mathbf{x}_0, \mathbf{c})$ is only available at the end of the image generation process. This sparsity of reward makes it challenging for the diffusion model to identify which actions during the denoising process positively impact the final alignment and reward them appropriately.
As a result, the diffusion model struggles to learn effective strategies and may even adopt unnecessary or incorrect ones.
The classic RL strategies, such as constructing additional rewards, are not suitable here because it is difficult to evaluate the noisy images during the denoising process.
This motivates us to develop new RL strategies for training diffusion models to mitigate the negative effects of sparse rewards.
For a comprehensive discussion on the challenge of sparse reward, we refer the readers to Appendix~\ref{sec:sparse_reward}. 

\subsection{Strategy 1: Backward Progressive Training}
The conventional training methods involve training the model across all timesteps of the denoising process from the beginning. However, due to the complexity and large noise present in the early timesteps, the training process can be unstable and inefficient, especially with sparse rewards. 
We hypothesize that focusing on the final timesteps, where the generated images are more coherent and less noisy, could provide a more stable foundation for the RL training. By mastering these final timesteps first, the model can incrementally handle the earlier, noisier stages more effectively, leading to overall better performance and control.
We call this strategy as backward progressive training (BPT).
Formally, let $T$ represent the total number of timesteps in the denoising process. Initially, we train the model on the last $\tau$ timesteps, where $\tau < T$.
Therefore, each trajectory sampled for training consists of $\tau$ timesteps:
\begin{equation}
\{s_t,a_t,\pi_\theta(a_t \mid s_t)|t=\tau,\tau-1,...,1\} \ \mathrm{with} \ \mathrm{reward} \ r(\mathbf{x}_0,\mathbf{c}),
\end{equation}
which can be abbreviated as $(s_{\tau:1},a_{\tau:1},\pi_{\tau:1},r)$ without ambiguity.
As training progresses, the training interval is extended backward by incorporating more timesteps, ultimately covering the entire range from $T$ to 1. The training objective during each phase remains consistent with Eq.~(\ref{loss}).
Following DDPO, we use policy gradient estimation~\cite{kakade2002approximately,schulman2017ppo} and the gradient is:
\begin{equation}
\scalebox{0.84}{
    $\nabla_{\theta} \mathcal{J}_{\text{BPT}} = -\mathbb{E} \left[  \sum_{t=1}^\tau \frac{p_{\theta}(\mathbf{x}_{t-1} \mid \mathbf{x}_t, \mathbf{c})}{p_{\theta_{\text{old}}}(\mathbf{x}_{t-1} \mid \mathbf{x}_t, \mathbf{c})} \nabla_{\theta} \log p_{\theta}(\mathbf{x}_{t-1} \mid \mathbf{x}_t, \mathbf{c}) \, \hat{r}(\mathbf{x}_0, \mathbf{c}) \right]$
    },
\label{eq:j_bpt}
\end{equation}
where $\theta_{\text{old}}$ is the parameters of diffusion model prior to update and $\hat{r}$ is the normalized value of reward $r$ (see Appendix~\ref{sec:implementation} for details). The expectation is taken over sampled denoising trajectories.

Previous works fine-tune diffusion models along the entire denoising process from $x_T$ to $x_0$, with the sparse reward $r_0$. With such sparse reward, it is difficult for the model to directly learn effective network parameters for the entire denoising process. We propose BPT to make the model learn the denoising process from $x_\tau$ to $x_0$ first. As training progresses, $\tau$ is gradually increased to $T$, and the model learns to manage the earlier timesteps after becoming proficient in later timesteps. This is easier than directly learning the entire denoising process. By applying BPT, the model can more effectively learn how to denoise when only $x_0$, state at the last timestep, has a reward. We refer the readers to Appendix~\ref{sec:sparse_reward} for a comprehensive discussion. 

\subsection{Strategy 2: Branch-based Sampling}
The sparse rewards make it difficult to tell whether actions on certain timesteps during denoising have a positive or negative effect on the final alignment. 
To further mitigate this issue, we introduce the strategy of branch-based sampling (BS). When constructing training data for RL, we perform branch sampling at the beginning of training interval $[\tau,1]$, as shown in Figure~\ref{fig3}~(b). Within each branch, we divide the sampled denoising trajectories (distinguished by the superscript $i$) into two groups: 
\begin{equation}
\begin{aligned}
\mathcal{G}_+ = \left\{ \left( s_{\tau:1}^i, a_{\tau:1}^i, \pi_{\tau:1}^i, \hat{r}^i \right) \big| \hat{r}^i := \hat{r}(\mathbf{x}^i_0, \mathbf{c}) > 0 \right\},\\
\mathcal{G}_- = \left\{ \left( s_{\tau:1}^i, a_{\tau:1}^i, \pi_{\tau:1}^i, \hat{r}^i \right) \big| \hat{r}^i := \hat{r}(\mathbf{x}^i_0, \mathbf{c}) < 0 \right\},
\end{aligned}
\end{equation}
where group $\mathcal{G}_+$ consists of trajectories with positive rewards (if available), and group $\mathcal{G}_-$ consists of trajectories with negative rewards (if available). We then select the trajectory $(s_{\tau:1}^+, a_{\tau:1}^+, \pi_{\tau:1}^+, \hat{r}^+)$ with the best reward from the positive group and the trajectory $(s_{\tau:1}^-, a_{\tau:1}^-, \pi_{\tau:1}^-, \hat{r}^-)$ with the worst reward from the negative group to form a contrastive sample pair for RL. The gradient of the contrastive sample pair is:
\begin{equation}
\scalebox{0.84}{
\begin{math}
\begin{aligned}
\nabla_{\theta} \mathcal{J}_{\text{BS}} = -\mathbb{E} \Bigg( \sum_{t=1}^{\tau} \Bigg[ &\frac{p_{\theta}(\mathbf{x}_{t-1}^{+} \mid \mathbf{x}_{t}^{+}, \mathbf{c})}{p_{\theta_{\text{old}}}(\mathbf{x}_{t-1}^{+} \mid \mathbf{x}_t^{+}, \mathbf{c})} \nabla_{\theta} \log p_{\theta}(\mathbf{x}_{t-1}^{+} \mid \mathbf{x}_t^{+}, \mathbf{c}) \hat{r}^{+} \\ 
+ &\frac{p_{\theta}(\mathbf{x}_{t-1}^{-} \mid \mathbf{x}_{t}^{-}, \mathbf{c})}{p_{\theta_{\text{old}}}(\mathbf{x}_{t-1}^{-} \mid \mathbf{x}_t^{-}, \mathbf{c})} \nabla_{\theta} \log p_{\theta}(\mathbf{x}_{t-1}^{-} \mid \mathbf{x}_t^{-}, \mathbf{c}) \hat{r}^{-}  \Bigg] \Bigg).
\end{aligned}
\end{math}
}
\label{eq:j_bs}
\end{equation}

By isolating the impact of actions outside the training interval on the final images, the comparison between the contrastive samples directly reflects how much the actions within the training interval contribute to the reward.
Branch-based sampling strategy provides clear signals to the model, allowing the model to focus on actions that truly drive positive outcomes. Therefore, it further mitigates the impact of reward sparsity and facilitates more efficient learning of effective policies. Moreover, by avoiding learning unnecessary policies (\textit{e.g.}, image styles), our approach can also maintain the diversity of generated images, which will be demonstrated and discussed in the following section.
We emphasize that $\text{B}^2\text{-DiffuRL}$ does not increase computational cost of RL algorithms, as discussed in Appendix~\ref{sec:computational_cost}.

%% file: Section/experiment.tex
\section{Experiments}

\begin{figure*}[t]
    % \vspace{-0.4em}
    \centering
    \includegraphics[width=1.0\linewidth]{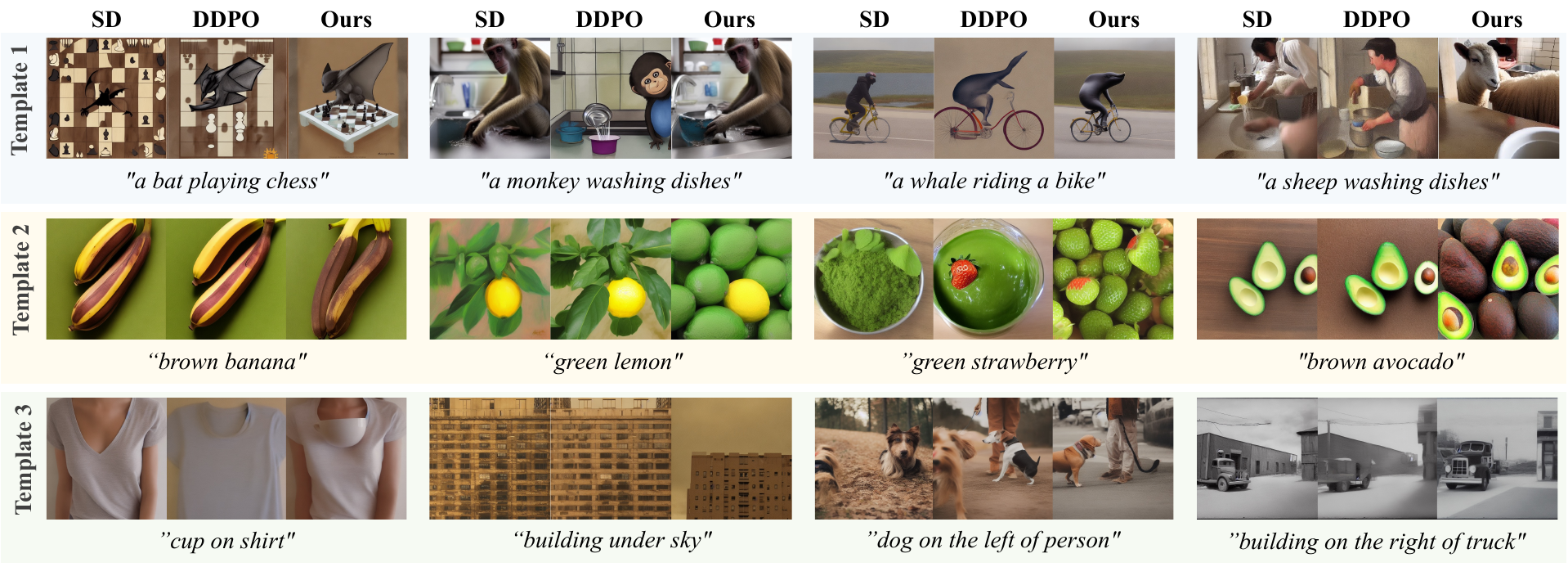}
    \caption{\textbf{(Samples)} Examples of images generated by different methods on three templates. For each set of images, we use the same random seed. Our method achieves better prompt-image alignment compared to vanilla Stable Diffusion and DDPO.}
    \label{examples}
    % \vspace{-0.4em}
\end{figure*}

In this section, we evaluate the effectiveness of $\text{B}^2\text{-DiffuRL}$ in terms of improving prompt-image alignment and maintaining diversity. We first compare our method with existing state-of-the-art method DDPO~\cite{Black2023TrainingDM}. 
Then, we focus on ablation studies on the proposed two strategies, as well as the compatibility and generalization ability. 
For simplicity, we refer to $\text{B}^2\text{-DiffuRL}$ as ours in this section.

\subsection{Experimental Setup}

\noindent\textbf{Diffusion Models.} Following the previous work~\cite{Black2023TrainingDM}, we use Stable Diffusion (SD) v1.4 as the backbone diffusion model, which has been widely used in academia and industry.
We apply LoRA to UNet for efficient fine-tuning~\cite{hu2021lora}. 
We employ DDIM~\cite{song2022denoising} algorithm for sampling. Following the previous work~\cite{Black2023TrainingDM}, we set the total denoising timesteps $T=20$. The weight of noise is set to 1.0, which decides the degree of randomness of each denoising in DDIM. Each experiment is conducted with three different seeds. 

\noindent\textbf{Prompt Templates.} In the sampling phase, we construct the prompts based on three different templates. 
The three prompt templates consider the behavior of the object, the attribute of the object, and the positional relationship between the objects in turn, which we believe can cover a wide range of commonly used prompts in image generation.
(1) Template 1:``\textit{a(n) [animal] [activity]}''. We use this template designed by DDPO. The animal is chosen from the list of 45 common animals, and randomly matched with an activity from the list:``\textit{riding a bike}'', ``\textit{playing chess}'' and ``\textit{washing dishes}''.
(2) Template 2: ``\textit{[color] [fruit/vegetable]}''. This template focuses on object attributes. 
To construct a list of color-fruit/vegetable combinations, we query GPT-4~\cite{achiam2023gpt} about fruits/vegetables' names and their common colors. We require each item to have at least 3 colors, and we end up building 40 prompts for this template.
(3) Template 3: ``\textit{[object 1] [predicate] [object 2]}''. The predicates refer to positional relationship. 
We construct the prompts based on the annotations of Visual Relation Dataset~\cite{lu2016visual}. We choose four predicates: ``\textit{on}'', ``\textit{under}'', ``\textit{on the left of}'', and ``\textit{on the right of}'', and end up with 40 prompts for this template.
The prompts mentioned above are only used for training. In order to evaluate the generalization ability, we further construct prompts that will not be used in training.
The full prompt lists are shown in the Appendix~\ref{sec:prompt_list}.

\noindent\textbf{Rewards.} We score the prompt-image alignment by BERTScore and CLIPScore, and use them as reward functions:
(1) BERTScore is introduced by DDPO~\cite{Black2023TrainingDM}, in which one uses the visual language model, such as LLaVA~\cite{liu2023llava}, to generate a description of the image, and then uses BERT's recall metric~\cite{devlin2018bert} to measure the semantic similarity between the prompt and the description.
(2) CLIPScore is simply the similarity between text embedding and image embedding measured by CLIP model~\cite{radford2021clip, bie2023renaissance}.
We recommend using CLIPScore as reward function due to the instability of BERTScore, as shown in Appendix~\ref{sec:compare_reward}. For implementations, we use 7b half-precision LLaVA v1.5 model~\cite{liu2023improved}, DeBERTa xlarge model~\cite{he2021deberta} (a variant of BERT model), and ViT-H-14 CLIP model~\cite{radford2021clip}, respectively.
To improve the stability of training, we normalize the rewards, as described in detail in Appendix~\ref{sec:implementation}. 

\noindent\textbf{Evaluation Metrics.} In this paper, we focus on both prompt-image alignment and image diversity.
For alignment, we use BERTScore~\cite{Black2023TrainingDM} and CLIPScore~\cite{radford2021clip, bie2023renaissance} as metrics, the same as reward functions. A higher BERTScore or CLIPScore represents better prompt-image alignment. 
For diversity, following previous works~\cite{barratt2018note, bie2023renaissance, ahn2019variational, zhao2020diffaugment}, we use inception score (IS) as the metric.
A higher inception score represents better image diversity. 

\begin{figure*}
    \begin{minipage}{0.7\linewidth}
        % \vspace{-0.4em}
        \centering
        \setlength{\abovecaptionskip}{0.25cm}
        \includegraphics[width=1.0\linewidth]{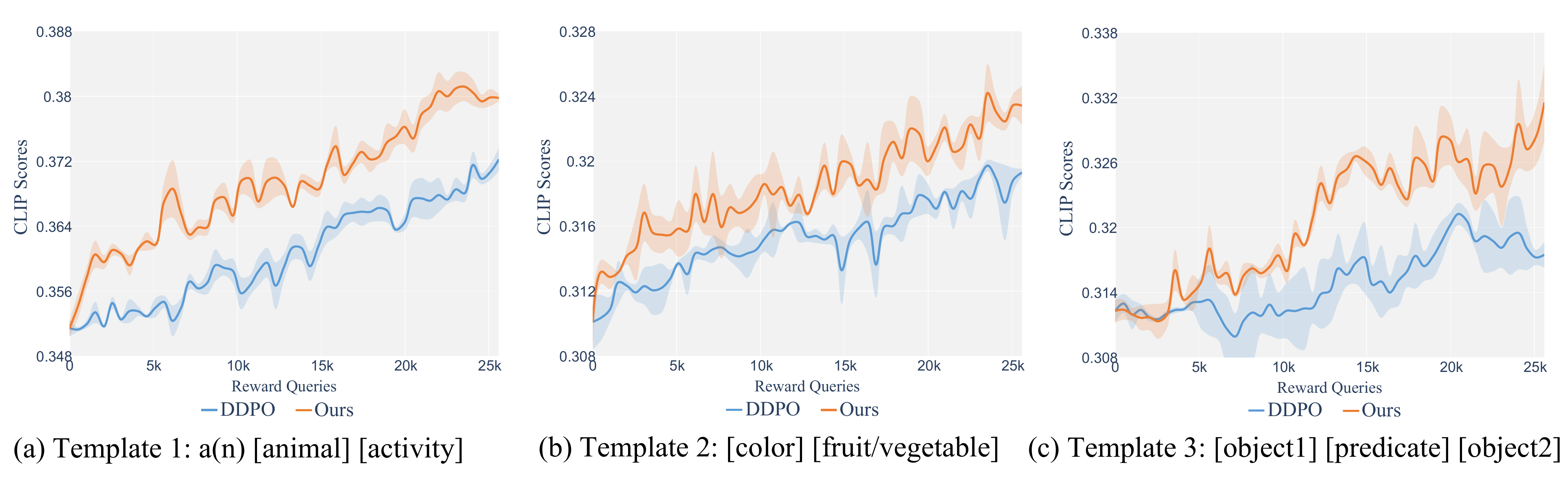}
        \caption{\textbf{(Alignment)} Alignment curves of our method and DDPO on three prompt templates.}
        \label{alignment}
    \end{minipage}
    \hfill
    \begin{minipage}{0.3\linewidth}
        \centering
        \setlength{\abovecaptionskip}{1.5em}
        \vspace{0.6em}
        \scalebox{0.8}{
        \begin{tabular}
        {c|ccc}
            \toprule
            Methods & Temp. 1 & Temp. 2 & Temp. 3  \\ \midrule
            SD & 1.3179 & 1.4133 & 1.3582 \\ 
            DDPO & 1.2886 & 1.3323 & 1.3273 \\ 
            Ours & \textbf{1.3127} & \textbf{1.3579} & \textbf{1.3348} \\ \bottomrule
        \end{tabular}
        }
        \captionof{table}{\textbf{(Diversity)} IS~$\uparrow$ of images generated by the SD~\cite{rombach2022latent}, DDPO~\cite{Black2023TrainingDM}, and ours on three templates. There is a trade-off between alignment and diversity, while our method helps maintain diversity.}
        \label{tab:template_diversity}
    \end{minipage}
    \vspace{0.3em}
\end{figure*}

\begin{figure*}
    \begin{minipage}{0.7\linewidth}
        \centering
        \includegraphics[width=1.0\linewidth]{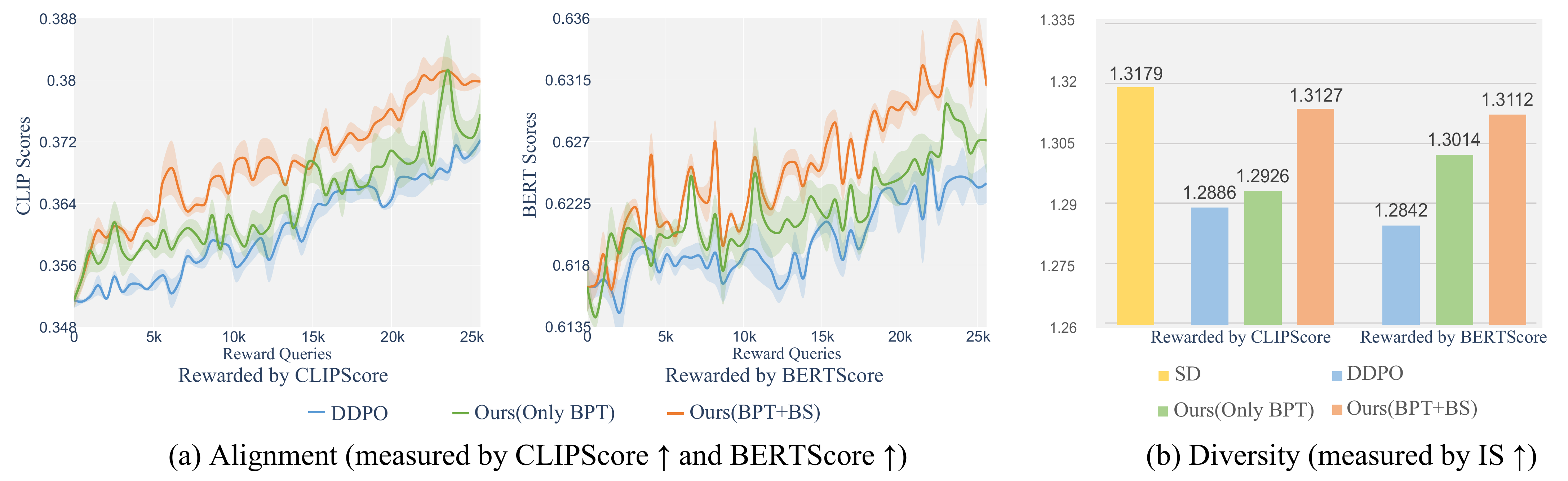}
        \caption{\textbf{(Ablation Study)} We separately evaluate the impact of each proposed strategy on prompt-image alignment and image diversity with template 1. (a) Both BPT and BS strategies help improve prompt-image alignment. (b) BS strategy also helps to maintain image diversity.}
        \label{ablation}
        % \vspace{-0.4em}
    \end{minipage}
    \hfill
    \begin{minipage}{0.28\linewidth}
        \centering
        \includegraphics[width=0.9\linewidth]{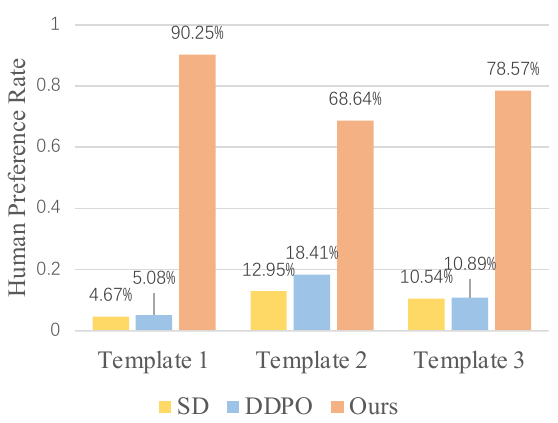}
        \caption{\textbf{(Human Evaluation)} Human preference rates for prompt-image alignment of images generated by SD, DDPO and our method.}
        \label{human_evaluation}
    \end{minipage}
\end{figure*}

\subsection{Qualitative Evaluation}

We first evaluate the performance of our method and DDPO on the three prompt templates rewarded by CLIPScore. 
We use our method and DDPO respectively to fine-tune the diffusion model. After the same round of training, we sample some images from original model and fine-tuned models, as shown in Figure~\ref{examples}.
The results qualitatively show that our method performs better than DDPO in improving the prompt-image alignment. 
We also conduct human preference test over 80 independent human raters (from undergrad to Ph.D.), who are asked to pick the best fit to prompt among three images generated by different models. As shown in Figure~\ref{human_evaluation}, the images generated by our method get higher preference rates than original SD and DDPO on all the three prompt templates.
Also, images by our method are more diverse than those by DDPO. For example, on template 1, all images by DDPO adopt a cartoon style, while those by ours keep original styles of SD; on templates 2 and 3, backgrounds of the images by DDPO tend to reduce to a single color, while those by ours do not. This can be seen more clearly from Appendix~\ref{sec:more_samples}.

\subsection{Quantitative Evaluation}

We compare our method with DDPO quantitatively in terms of prompt-image alignment and diversity. 

\noindent\textbf{Prompt-Image Alignment.} Figure~\ref{alignment} shows the curve of CLIPScore when fine-tuning the diffusion models using our method and DDPO as the amount of reward queries increases.
We can observe that our method almost always achieves higher CLIPScore during fine-tuning on all the three prompts. 
This shows that our approach can improve prompt-image alignment better with the same number of reward queries compared to DDPO, which is due to our proposed two strategies.

\noindent\textbf{Image Diversity.} We evaluate the diversity of the images generated by original SD and the models fine-tuned by our method and DDPO. 
The results are shown in Table~\ref{tab:template_diversity}.
After 25.6k reward queries during fine-tuning, both the models trained by ours and DDPO exhibit a reduction in diversity, since there is an inherent trade-off between alignment and diversity~\cite{zhang2024inherent}.
However, we find that the models trained by our method have a smaller reduction in diversity on all templates. 
For example, on template 1, the diversity of the model trained by our method decreases much less than that of DDPO, and is basically the same as the original model.
Overall, our method can mitigate the reduction in image diversity during RL-based diffusion model fine-tuning. 

\begin{figure*}
    \begin{minipage}{0.7\linewidth}
        % \vspace{-0.4em}
        \centering
        \setlength{\abovecaptionskip}{0.cm}
        \includegraphics[width=1.0\linewidth]{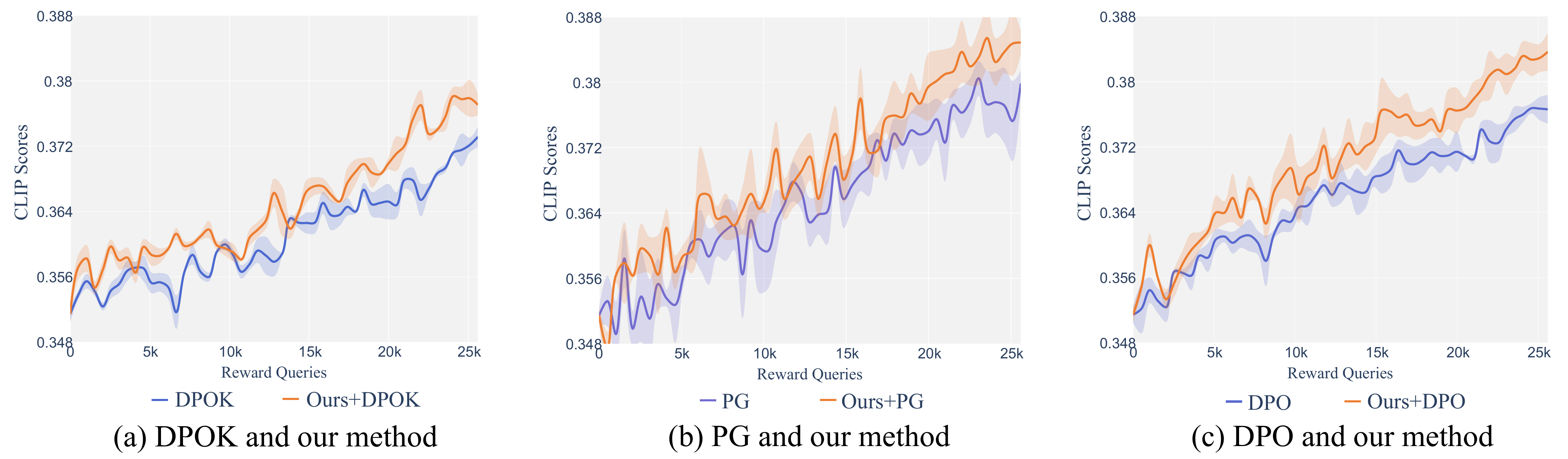}
        \caption{\textbf{(Compatibility: Alignment)} Alignment curves of our method on template 1 when compatible with difference RL algorithms. }
        \label{compatibility_alignment}
        % \vspace{-0.4em}
    \end{minipage}
    \hfill
    \begin{minipage}{0.28\linewidth}
        \centering
        \small
        \setlength{\abovecaptionskip}{1.em}
        \vspace{0.4em}
        \begin{tabular}{c|cc}
            \toprule
            Methods & Vanilla & Ours+Vanilla \\ \midrule
            SD & 1.3179 & - \\
            DPOK & 1.2785 & \textbf{1.3005} \\
            PG & 1.2462 & \textbf{1.2896} \\
            DPO & 1.2895 & \textbf{1.3051} \\ \bottomrule
        \end{tabular}
        \captionof{table}{\textbf{(Compatibility: Diversity)} IS~$\uparrow$ of images on template 1 generated by our method when compatible with different RL algorithms.}
        \label{compatibility_diversity}
    \end{minipage}
\end{figure*}

\begin{figure*}
    \begin{minipage}{0.64\linewidth}
        \centering
        \setlength{\abovecaptionskip}{0.em}
        \includegraphics[width=0.96\linewidth]{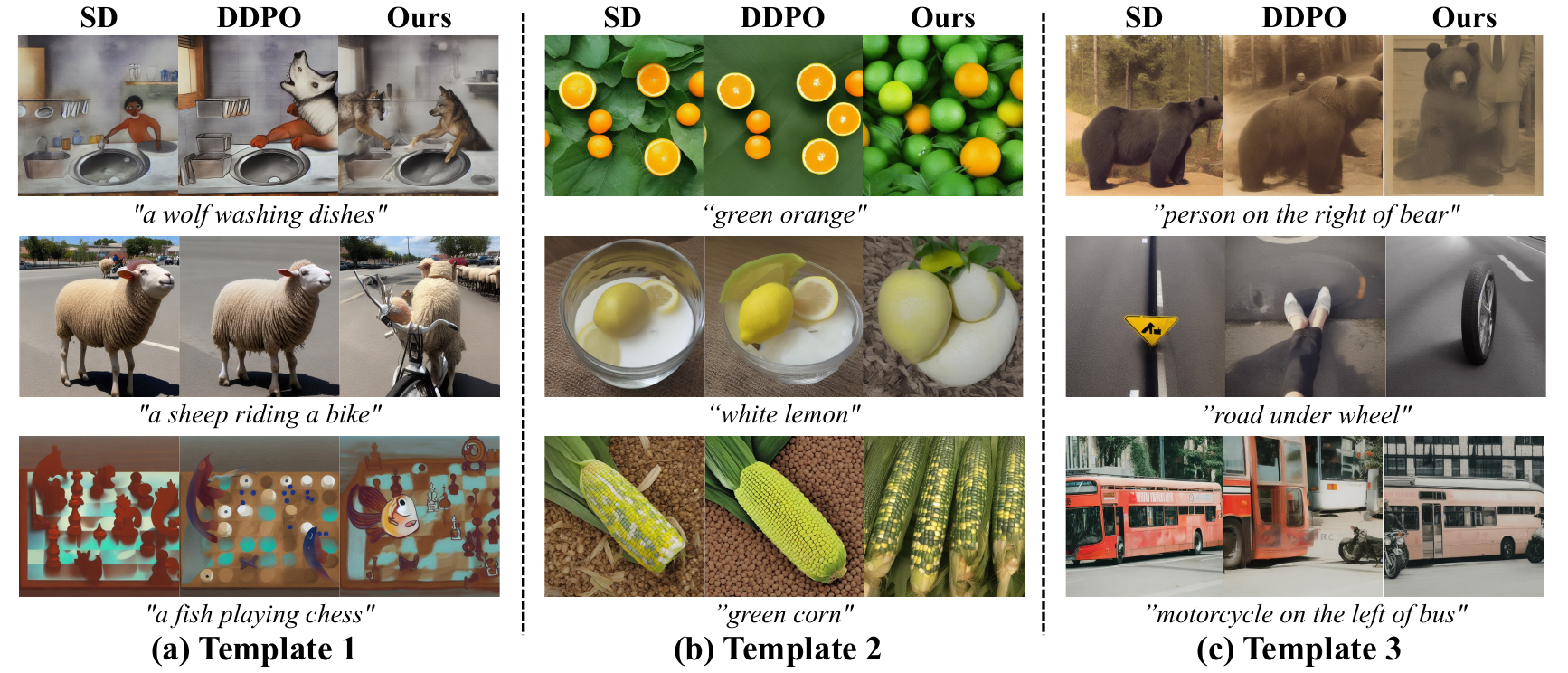}
        \caption{\textbf{(Generalization)} Examples of images generated by SD, DDPO and ours on three templates. The prompts are not used in training. We use the same seed for sampling.}
        \label{generalization_example}
        % \vspace{-0.4em}
    \end{minipage}
    \hfill
    \begin{minipage}{0.33\linewidth}
        \centering
        \small
        % \setlength{\abovecaptionskip}{2.em}
        % \vspace{0.em}
        \begin{tabular}
        {c|ccc}
            \toprule
            Methods & Temp. 1 & Temp. 2 & Temp. 3 \\ \midrule
            SD & 0.3515 & 0.3168 & 0.2977 \\ 
            DDPO & 0.3698 & 0.3175 & 0.3134 \\
            Ours & \textbf{0.3748} & \textbf{0.3252} & \textbf{0.3183} \\ \bottomrule
        \end{tabular}
        \captionof{table}{\textbf{(Generalization)} Prompt-image alignment (measured by CLIPScore~$\uparrow$) of the generated images by SD, DDPO and our method on the prompts based on three templates. The prompts are not used during the training process.}
        \label{tab:template_generalization}
    \end{minipage}
\end{figure*}

\subsection{Ablation Study}

We separately evaluate the impact of each proposed strategy on alignment and diversity respectively.

\noindent\textbf{Ablation Study on BPT Strategy.}
To evaluate the effectiveness of BPT, we fine-tune Stable Diffusion with only BPT strategy, rewarded by CLIPScore and BERTScore respectively. As shown in Figure~\ref{ablation}~(a), regardless of the reward function, our proposed BPT strategy outperforms DDPO in terms of alignment.
As we previously analyzed, BPT simplifies learning by training in stages, alleviating the negative effects of sparse rewards, and thus improving alignment.
Moreover, since we only train models on timesteps of current training interval instead of all the denoising process, the computation costs of our method are less than DDPO for each queried reward.

\noindent\textbf{Ablation Study on BS Strategy.}
The effectiveness of BS strategy on prompt-image alignment is shown in Figure~\ref{ablation}~(a). We can observe that, based on BPT, the BS strategy further improves alignment in terms of both BERTScore and CLIPScore. By comparing contrastive samples, BS provides a clear indication of how much the policies of current training interval contribute to final images. This helps the model to learn effective policies.
Additionally, as shown in Figure~\ref{ablation}~(b), diversity of generated images always suffers from reduction since there is a trade-off between alignment and diversity. Fortunately, the BS strategy helps models avoid learning unnecessary policies, thereby contributing to maintaining image diversity. With BS strategy, diversity of the fine-tuned model decreases less, and even achieves similar diversity as the original SD.

\subsection{Compatibility}

Our framework $\text{B}^2\text{-DiffuRL}$ is compatible with various RL algorithms, not limited to DDPO. We further apply $\text{B}^2\text{-DiffuRL}$ to some widely used RL algorithms in diffusion model fine-tuning, including DPOK~\cite{fan2023dpok}, policy gradient (PG)~\cite{schulman2017ppo} and direct preference optimization (DPO)~\cite{rafailov2023direct, wallace2023diffusion}. The implementation details are shown in Appendix~\ref{sec:implementation}. 
On the one hand, as we can see from Figure~\ref{compatibility_alignment}, when compatible with different RL algorithms, our method can help each of them to improve alignment to a greater extent. 
On the other hand, as shown in Table~\ref{compatibility_diversity}, while all algorithms reduce the diversity of generated images, our method can help mitigate the reduction. 
These experimental results further illustrate the effectiveness of our method in terms of both prompt-image alignment and diversity when applied to various RL algorithms. 

\subsection{Generalization Ability}

Models fine-tuned by our method show generalization capabilities. 
We generate 1,600 images on the prompts based on the corresponding templates but not belong to the training lists, and test the prompt-image alignment on CLIPScore. 
As shown in Table~\ref{tab:template_generalization}, compared with DDPO, the models fine-tuned with our method also perform better on these prompts not used for training.
Figure~\ref{generalization_example} shows examples of images generated on these prompts, qualitatively illustrating the good generalization ability of the models fine-tuned with our method.
More samples can be seen in Appendix~\ref{sec:more_samples}.

% \subsection{Discussion}

%% file: Section/conclusion.tex
\section{Conclusions}

In this work, we mitigated the issues of prompt-image misalignment in text-to-image diffusion models by reinforcement learning (RL). 
We highlight the challenge of sparse reward when training diffusion models with RL.
By introducing a compatible RL-based fine-tuning framework $\text{B}^2\text{-DiffuRL}$ that leverages backward progressive training and branch-based sampling strategies, we effectively mitigated the negative effects of sparse reward.
Using Stable Diffusion as backbone, we performed extensive experiments with various kinds of text prompts.
Both qualitative and quantitative experimental results demonstrate that, compared with naive RL-based diffusion model training method, the proposed framework achieves better prompt-image alignment while sacrificing less image diversity.

%% file: Section/appendix.tex
\noindent The \textbf{Appendix} is organized as follows:
\begin{itemize}[leftmargin=*]
    \item \textbf{Appendix~\ref{sec:broad_impacts}:} discusses the potential broader impacts of our work. 
    \item \textbf{Appendix~\ref{sec:symbol_table}:} gives the list of abbrevations and symbols in our paper. 
    \item \textbf{Appendix~\ref{sec:sparse_reward}:} gives a comprehensive discussion on the challenge of sparse reward. 
    \item \textbf{Appendix~\ref{sec:implementation_detail}:} provides more details on implementation (\textit{e.g.}, experimental resources and hyperpatameters). 
    \item \textbf{Appendix~\ref{sec:pseudo_code}:} provides pseudo-code of $\text{B}^2\text{-DiffuRL}$. 
    \item \textbf{Appendix~\ref{sec:eval_metrics}:} gives an discussion on evaluation metrics, including comparison between BERTScore and CLIPScore, and inception score.
    \item \textbf{Appendix~\ref{sec:more_samples}:} provides more image samples generated by the diffusion models fine-tuned with $\text{B}^2\text{-DiffuRL}$.
    \item \textbf{Appendix~\ref{sec:prompt_list}:} provides the prompt lists used in our experiments.
\end{itemize}

\section{Broader Impacts}
\label{sec:broad_impacts}
Generative models, particularly diffusion models, are powerful productivity tools with significant potential for positive applications. However, their misuse can lead to undesirable consequences. Our research focuses on improving the prompt-image alignment of diffusion models, enhancing their accuracy and usefulness in fields such as medical image synthesis. While these advancements have clear benefits, they also pose risks, including the creation of false information that can mislead the public and manipulate public opinion. Therefore, ensuring reliable detection of synthesized content is crucial to mitigate the potential harm associated with generative models.

\section{Abbreviation and Symbol Table}
\label{sec:symbol_table}

The list of important abbreviations and symbols in this paper goes as Table~\ref{tab:symbol}. 

\begin{table*}[h]
    \centering
    \begin{tabular}
    {ll}
        \toprule
        Abbreviation/Symbol & Meaning \\ 
        \midrule
        \multicolumn{2}{c}{\textit{\underline{Abbreviations of Concepts}}} \\
        DM & Diffusion Model \\
        RL & Reinforcement Learning\\
        SD & Stable Diffusion \\
        LoRA & Low-Rank Adaptation \\
        DDIM & Denoising Diffusion Implicit Model \\
        CLIP & Contrastive Language-Image Pre-Training \\
        BERT & Bidirectional Encoder Representation from Transformers \\
        IS & Inception Score \\
        \midrule
        \multicolumn{2}{c}{\textit{\underline{Abbreviations of Approaches}}} \\
        $\text{B}^2\text{-DiffuRL}$ & BPT and BS for Reinforcement Learning in Diffusion models \\
        BPT & Backward Progressive Training \\
        BS & Branch-based Sampling \\
        DDPO & Denoising Diffusion Policy Optimization \\
        DPOK & Diffusion Policy Optimization with KL regularization \\
        PG & Policy Gradient algorithm \\
        DPO & Direct Preference Optimization \\
        \midrule
        \multicolumn{2}{c}{\textit{\underline{Symbols of Diffusion Models}}} \\
        $x_0$ & Generated image \\
        $x_t$ & Image with noise at timestep $t$\\
        $c$ & Condition for image generation, also called prompt \\
        $\theta$ & Parameters of the diffusion model \\
        $\mu_{\theta}, \Sigma_{\theta}$ & Mean and variance predicted by the diffusion model \\
        $\mathcal{N}()$ & Gaussian distribution \\
        $T$ & Total timesteps \\
        $[\tau, 1]$ & Training interval from timestep $\tau$ to $1$ \\
        \midrule
        \multicolumn{2}{c}{\textit{\underline{Symbols of Reinforcement Learning}}} \\
        $s_t$ & State at timestep $t$ \\
        $a_t$ & Action at timestep $t$ \\
        $\pi_\theta$ & Action selection policy parameterized by $\theta$ \\
        $r()$ & Reward function \\
        $\hat{r}()$ & Reward function with normalization \\
        \bottomrule
    \end{tabular}
    \caption{List of important abbreviations and symbols.}
    \label{tab:symbol}
\end{table*}

\section{A Comprehensive Discussion on Sparse Rewards}
\label{sec:sparse_reward}

\begin{figure}[h]
    \vspace{-1.4em}
    \centering
    \setlength{\abovecaptionskip}{0.em}
    \includegraphics[width=\linewidth]{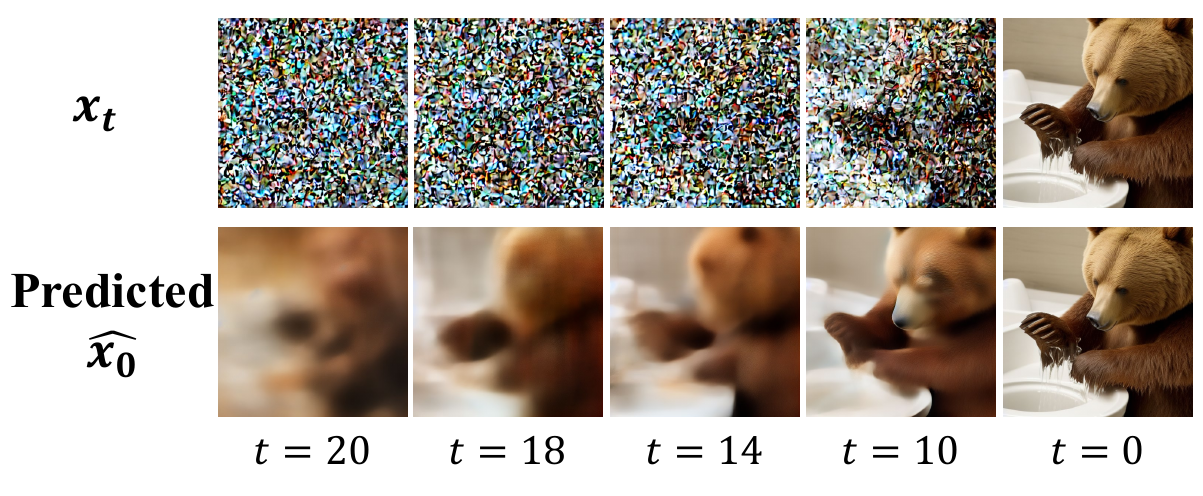}
    \caption{\textbf{(Examples for Predicted $\hat{x}_0$)} This figure shows the $x_t$ and predicted $\hat{x}_0$ in the denoising process.}
    \label{sparse_fig3}
    \vspace{-0.6em}
\end{figure}

\begin{figure*}[h]
    \centering
    \includegraphics[width=1.0\linewidth]{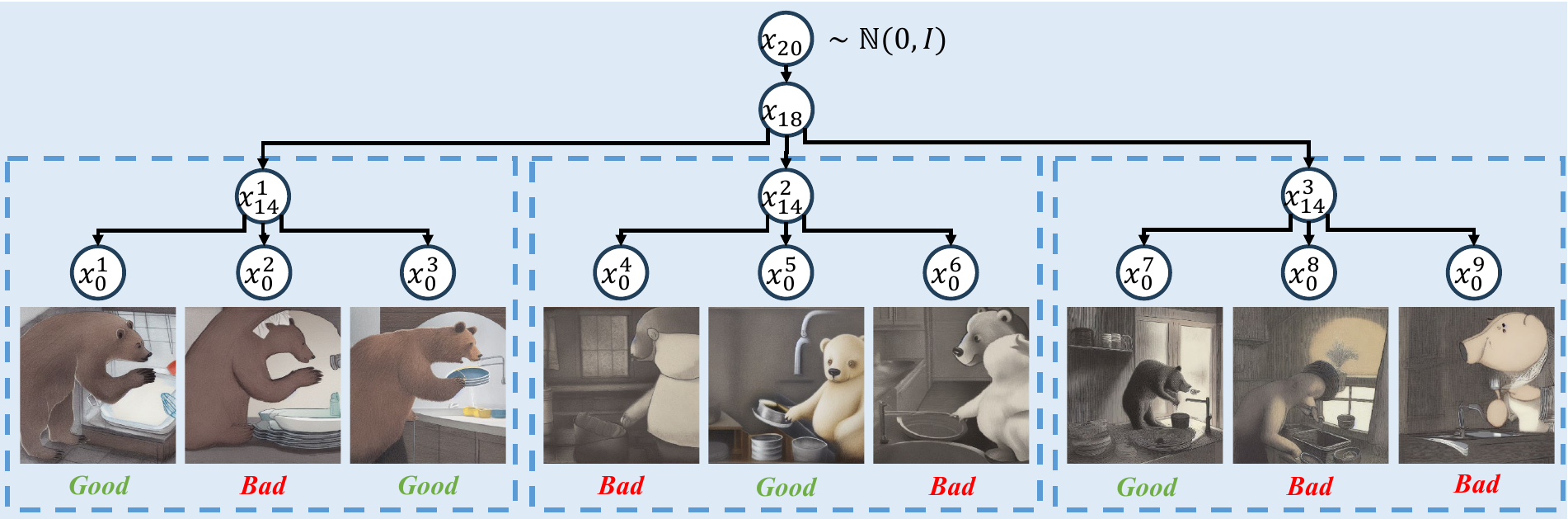}
    \caption{\textbf{(Examples Showing the Problem of Sparse Reward)} For these examples, the number of denoising timesteps $T$ is set to 20, and the prompt is \textit{``a bear washing dishes''}. The images are denoised from the same $x_{18}$ with different seeds, and every 3 images are denoised from the same $x_{14}$ with different seeds. As we can see, the images denoised from the same parent node $x_{18}$ or $x_{14}$ can get different alignment scores. We can not tell whether $x_{18}$/$x_{14}$ is good or bad from one final image. Consequently, it is inappropriate to use the reward for the last timesteps as the reward for the whole denoising process. }
    \label{sparse_fig1}
    \vspace{0.5em}
\end{figure*}

\noindent\textbf{(1) \textit{How does the sparse reward make a negative impact on RL-based diffusion models fine-tuning?}} The reward is sparse when we execute RL-based diffusion models fine-tuning, since only the final image $x_0$ is available to evaluate the text-image alignment. Previous works such as DDPO and DPOK have to treat the denoising actions at different timesteps equally and set $r_{T-1}=r_{T-2}=...=r_0$. However, we argue that the denoising actions $a_t$ on different timesteps have different effects on alignment, and the unreasonable reward setting is not conducive to learning. For example, as shown in Figure~\ref{sparse_fig1}, the images $x^1_0$, $x^2_0$, and $x^3_0$ have the same parent node $x^1_{14}$ but different text-image alignment scores. The reason for their difference is that different denoising actions $a_{13:1}$ (instead of $a_{20:14}$). Therefore, it is inappropriate to use sparse reward $r_0$ to reward denoising actions $a_{20:14}$. Besides, as shown in Table~\ref{sparse_tab1}, the differences in alignment results under the same branch are common, even with a small number of timesteps $T=20$. This reveals the universality of the sparse reward problem.

\noindent\textbf{(2) \textit{Why not directly calculate the alignment score of the predicted $\hat{x}_0$ at each timestep $t$?}} 
Each DDPM or DDIM denoising step can generate a corresponding predicted $\hat{x}_0$ using $x_t$ and the predicted $\epsilon$. 
However, as shown in Figure~\ref{sparse_fig3}, the predicted $\hat{x}_0$ at most denoising steps is unclear. We do not think that the reward function for final images can make an accurate evaluation of intermediate images.

\noindent\textbf{(3) \textit{How do the proposed BPT and BS strategies help to mitigate the sparse reward issue?}} BPT allows diffusion models to focus on specific training intervals (from $\tau$ to $0$) rather than all timesteps (from $T$ to $0$). 
% As training interval extends, reward will eventually be assigned to $a_{T:\tau+1}$. 
As training progresses, $a_{\tau:1}$ turn to align better, thus the alignment is more determined by $a_{T:\tau+1}$, and the final reward is more accurate for $a_{T:\tau+1}$. 
That is, BPT helps to assign more appropriate rewards to denoising actions $a_{T:\tau+1}$. 
BS samples different images from the same parent node $x_\tau$ and selects the best one and the worst one to form a contrastive sample pair. By comparing the contrastive sample pair, BS can provide more accurate rewards for denoising actions $a_{\tau:1}$, since the images within the same branch have the same state $s_{\tau}$. Moreover, since the contrastive samples share high-level visual semantics such as image style, the models do not learn to generate images with a specific style. This is why our proposed strategies preserve higher diversity compared to naive RL algorithms.

\section{Implementation Details}
\label{sec:implementation_detail}

\subsection{Implementation of Our Method}
\label{sec:implementation}

\noindent \textbf{Proximal Policy Optimization.} Following DDPO, we apply proximal policy optimization (PPO) algorithm~\cite{schulman2017ppo}, a commonly used family of policy gradient (PG) algorithm for reinforcement learning. And we perform importance sampling $\frac{p_\theta(x_{t-1} | x_t, c) }{p_{\theta_{old}}(x_{t-1} | x_t, c)}$ and clipping~\cite{schulman2017ppo} to implement PPO. 

\noindent \textbf{Extendence of Training Interval.} When employing backward progressive training, the training interval will extend gradually to cover all timesteps of the denoising process. 
In practice, we use a linear expansion strategy. 
That is, given the initial training interval $[\tau_0, 1]$, the total timesteps $T$ and total number of training round $N$, the training interval in round $n$ is $[\tau_0+\lfloor \frac{T-\tau_0+1}{N} \rfloor, 1]$. 

\noindent \textbf{Reward Normalization.} The prompt-image alignment scores given by CLIP or BERT need to be normalized before being used as rewards in training. 
In practice, we compute the mean and variance of the scores for each training round, with the images generated by the same prompt in the current round and in the past several rounds. 
Then the score can be normalized as $\frac{score-mean}{variance}$. 
When computing mean and variance, we incorporate images from the past rounds into calculation, because calculation using only images from one single round may be inaccurate.
However, we only use images from the past few rounds, instead of all rounds, in the consideration that the scores of images multiple rounds ago differ greatly from those in current round as fine-tuning progresses, and are not suitable for estimating mean and variance of current round. 
In practice, we use images from the past $8$ training rounds. 

\noindent \textbf{Compatibility with Policy Gradient.} When applying PG, the value function $V(x_{\tau}, c)$ should be considered. In our implementation, we replace value function with the reward normalization mentioned above. 
What's more, the importance sampling $\frac{p_\theta(x_{t-1} | x_t, c) }{p_{\theta_{old}}(x_{t-1} | x_t, c)}$ is also applied to improve stability of training. Therefore, the optimization objective of PG in our setting is the same as Eq.~(\ref{eq:j_bs}), but without using the clipping in PPO. 

\begin{table*}[h]
    \centering
    \resizebox{1.0\linewidth}{!}{
    \begin{tabular}
    {c|cccccccccc}
        \toprule
        Timestep when Branching & 2 & 4 & 6 & 8 & 10 & 12 & 14 & 16 & 18 & 20 \\ 
        \midrule
        Propotion & 8.2\% & 16.8\% & 15.6\% & 28.1\% & 28.9\% & 34.0\% & 43.0\% & 44.5\% & 52.3\% & 66.4\% \\
        \bottomrule
    \end{tabular}
    }
    \caption{\textbf{(Proportion of branches that contain both well-aligned and poorly-aligned images when branching from different timesteps)} The number of denoising timesteps $T$ is set to 20. We sample 256 branches each time, and each branch contains 3 images. The differences in alignment results under the same branch are widespread.}
    \label{sparse_tab1}
\end{table*}

\noindent \textbf{Compatibility with DPOK.} DPOK also uses value function and clipping in their implementation. Same with PG, we replace value function with reward normalization. Therefore, the gradient of our method when compatible with DPOK goes as Eq.~(\ref{eq:dpok}).
\begin{equation}
\begin{aligned}
\mathbb{E} \Big( \sum_{t=1}^{\tau} \Big[ &-\alpha \nabla_{\theta} \log p_{\theta}(\mathbf{x}_{t-1}^{+} \mid \mathbf{x}_t^{+}, \mathbf{c}) \hat{r}^{+} \\
&+ \beta \nabla_{\theta}\text{KL}(p_{\theta}(\mathbf{x}_{t-1}^{+} \mid \mathbf{x}_t^{+}, \mathbf{c}) || p_{\theta_{\text{old}}}(\mathbf{x}_{t-1}^{+} \mid \mathbf{x}_t^{+}, \mathbf{c})) \\
&- \alpha \nabla_{\theta} \log p_{\theta}(\mathbf{x}_{t-1}^{-} \mid \mathbf{x}_t^{-}, \mathbf{c}) \hat{r}^{-}  \\
&+ \beta \nabla_{\theta}\text{KL}(p_{\theta}(\mathbf{x}_{t-1}^{-} \mid \mathbf{x}_t^{-}, \mathbf{c}) || p_{\theta_{\text{old}}}(\mathbf{x}_{t-1}^{-} \mid \mathbf{x}_t^{-}, \mathbf{c}))  \Big] \Big).
\end{aligned}
\label{eq:dpok}
\end{equation}

\noindent \textbf{Compatibility with Direct Preference Optimization.} In contrastive sample pairs, the positive samples are more preferred than negative samples. Therefore, we can apply direct preference optimization (DPO). The gradient of our method when compatible with DPO goes as Eq.~(\ref{eq:dpo}).
\begin{equation}
\begin{aligned}
-\mathbb{E} \Big( \sum_{t=1}^{\tau} \Big[ &\frac{p_{\theta}(\mathbf{x}_{t-1}^{+} \mid \mathbf{x}_{t}^{+}, \mathbf{c})}{p_{\theta_{\text{old}}}(\mathbf{x}_{t-1}^{+} \mid \mathbf{x}_t^{+}, \mathbf{c})} \nabla_{\theta} \log p_{\theta}(\mathbf{x}_{t-1}^{+} \mid \mathbf{x}_t^{+}, \mathbf{c}) \\
- &\frac{p_{\theta}(\mathbf{x}_{t-1}^{-} \mid \mathbf{x}_{t}^{-}, \mathbf{c})}{p_{\theta_{\text{old}}}(\mathbf{x}_{t-1}^{-} \mid \mathbf{x}_t^{-}, \mathbf{c})} \nabla_{\theta} \log p_{\theta}(\mathbf{x}_{t-1}^{-} \mid \mathbf{x}_t^{-}, \mathbf{c})   \Big] \Big).
\end{aligned}
\label{eq:dpo}
\end{equation}

\subsection{Discussion on Value Function}

A value function $V(x_t, c)$ is usually used in policy gradient training. By subtracting $r(x_0, c)$ with $V(x_t, c)$, the variance of gradient estimation can be minimized~\cite{fan2023dpok}. 
However, we do not employ value function in our implementation. 

Branch-based sampling and reward normalization have the same effect as value function.
Since value function is trained to minimize $E_{p_\theta(x_{0:t})}(r(x_0, c)-V(x_t, c))^2$, the state value $V(x_t, c)$ approximately equals to the mean score of the $x_0$s denoised from the given $x_t$. In our approach, reward normalization, as detailed in Appendix~\ref{sec:implementation}, normalizes the score/reward using $\frac{score-mean}{variance}$, similar to the effect of applying value function (\textit{i.e.}, $(r(x_0, c)-V(x_{\tau}, c))$. Simultaneously, branch-based sampling provides additional samples denoised from the given $x_\tau$, which improves the estimation of the mean score. Moreover, the contrastive samples are denoised from the same $x_\tau$, with differences in their rewards reflecting variations in the denoising process from $x_\tau$ to $x_0$. By constructing pair-wise contrastive samples, branch-based sampling (BS) introduces reward signals that are independent of previous timesteps, helping to estimate the reward of $x_\tau$ accurately. This is also why applying BPT+BS consistently outperforms only applying BPT in our experiments.

\clearpage
\begin{algorithm*}
    \caption{Pseudo-code of $\text{B}^2\text{-DiffuRL}$ for one training round. }
    \label{alg:pseudo_code}
    \SetAlgoLined
    \SetKwInOut{Input}{Input}\SetKwInOut{Output}{Output}
 
    \Input{Denoising timesteps $T$, inner epoch $E$, number of samples each round $N$, prompt list $C$, number of branches $K$, training interval $[\tau, 1]$, reward function $r$, pretrained diffusion model $p_\theta$}.
    
    $p_{old} = \text{deepcopy}(p_\theta)$ \;
    $p_{old}.\text{require\_grad}($False$)$ \;
    \tcp{Sampling}
    $D_{sampling} = \{\}$ \;
    \For{$n \leftarrow 1$ \KwTo $N$}{
        Randomly choose a prompt $c$ from $C$ \;
        Randomly choose $x_T$ from $\mathcal{N}(0, I)$ \;
        $x_{(T-1):\tau} =$ Denoise with $p_\theta$ for $(T-\tau)$ steps \;
        \For{$k \leftarrow 1$ \KwTo $K$}{
            $x_{\tau}^k = \text{deepcopy}(x_{\tau})$ \;
            $x_{(\tau-1):0}^k = $Denoise with $p_\theta$ for $\tau$ steps \;
        }
        $D_{sampling}.\text{push}([x_{\tau:0}^{1:K}, c])$ \;
    }
    \tcp{Evaluation}
    $D_{training} = \{\}$ \;
    \For{$[x_{\tau:0}^{1:K}, c] \in D_{sampling}$}{
        $s^{1:K} = \text{normalization}(r(x_0^{1:K}, c))$  \tcp{Do normalization as Appendix~\ref{sec:implementation}}
        \eIf{$s^{1:K}$ contains both negative and positive scores}{
            $i = \text{argmax}(s^{1:K}); j = \text{argmin}(s^{1:K})$ \;
            $D_{training}.\text{push}([x_t^i, x_{t-1}^i, s^i, x_t^j, x_{t-1}^j, s^j, c]_{t=1:\tau})$  \tcp{Contrastive sample pairs}
        }{
            $i = \text{argmax}(abs(s^{1:K}))$ \;
            $D_{training}.\text{push}([x_t^i, x_{t-1}^i, s^i, c]_{t=1:\tau})$ \tcp{Simple samples}
        }
    }
    \tcp{Training}
    \For{$e \leftarrow 1$ \KwTo $E$}{
        $D = \text{shuffle}(D_{training})$ \;
        with grad \;
        \For {$d \in D$}{
            $d = \text{shuffle}(d)$ \;
            \eIf {$d$ is a contrastive sample pair}{
                \For{$[x_t^i, x_{t-1}^i, s^i, x_t^j, x_{t-1}^j, s^j, c] \in d$}{
                    update $\theta$ with gradient descent using Eq.~(\ref{eq:j_bs}) \;
                }
            }{
                \For{$[x_t^i, x_{t-1}^i, s^i, c] \in d$}{
                    update $\theta$ with gradient descent using Eq.~(\ref{eq:j_bpt}) \;
                }
            }
        }
    }
\end{algorithm*}
\clearpage

\subsection{Computational Cost}
\label{sec:computational_cost}

In experiments, it takes about 36 hours to reach 50 epochs using $\text{B}^2\text{-DiffuRL}$, while DDPO takes about 60 hours. Computational cost mainly consists of two parts: sampling and training. In training, for a sample $x_0$, the vanilla training using RL algorithm needs to traverse the entire denoising process from $x_T$ to $x_0$, while the training using BPT only needs to traverse from $x_\tau$ to $x_0$, where $\tau \leq T$. Therefore, using BPT leads to lower training cost in training. As for sampling, using branch-based sampling (BS) indeed leads to higher computational cost. However, sampling is much faster than training, so $\text{B}^2\text{-DiffuRL}$ requires lower computational cost overall.

\subsection{Experimental Resources}

We conducted experiments on 8 24GB NVIDIA 3090 GPUs. 
It took approximately 36 hours to reach 25.6k reward queries when rewarded by CLIPScore, and approximately 80 hours when rewarded by BERTScore (LLaVA inference would take much time). 

\subsection{Hyperparameters}

We list hyperparameters of our experiments in Table~\ref{tab:hyperparameters}. 

\begin{table}[h]
    \centering
    \scalebox{0.85}{
    \begin{tabular}
    {c|l|c|c}
        \toprule
        & Hyperpatameter & $\text{B}^2\text{-DiffuRL}$ & DDPO \\ 
        \midrule
        \multirow{6}*{Sampling} & Denoising steps $T$ & 20 & 20 \\
        & Noise Weight $\eta$ & 1.0 & 1.0 \\
        & Guidance Scale & 5.0 & 5.0 \\
        & Batch size & 8 & 8 \\
        & Batch count & 32 & 32 \\
        & Number of Branches & 3 & - \\
        \midrule
        \multirow{5}*{Optimizer} & Optimizer & AdamW~\cite{loshchilov2019decoupled} & AdamW \\ % \cline{2-4}
        & Learning rate & 1e-4 & 1e-4 \\
        & Weight decay & 1e-4 & 1e-4 \\
        & $(\beta_1, \beta_2)$ & (0.9, 0.999) & (0.9, 0.999) \\
        & $\epsilon$ & 1e-8 & 1e-8 \\
        \midrule
        \multirow{4}*{Training} & Batch size & 2 & 2 \\
        & Grad. accum. steps & 32 & 128 \\
        & Initial training interval & $[14, 1]$ & - \\
        & Score threshold & 0.5 & - \\
        \bottomrule
    \end{tabular}
    }
    \caption{Hyperparameters of our experiments.}
    \label{tab:hyperparameters}
\end{table}

\section{Pseudo-code}
\label{sec:pseudo_code}

The pseudo-code of $\text{B}^2\text{-DiffuRL}$ for one training round goes as Algorithm~\ref{alg:pseudo_code}. 

\section{Discussion on Evaluation Metrics}
\label{sec:eval_metrics}

\subsection{Comparison between BERTScore and CLIPScore}
\label{sec:compare_reward}

\begin{figure}[h]
    % \vspace{-0.4em}
    \centering
    \includegraphics[width=1.0\linewidth]{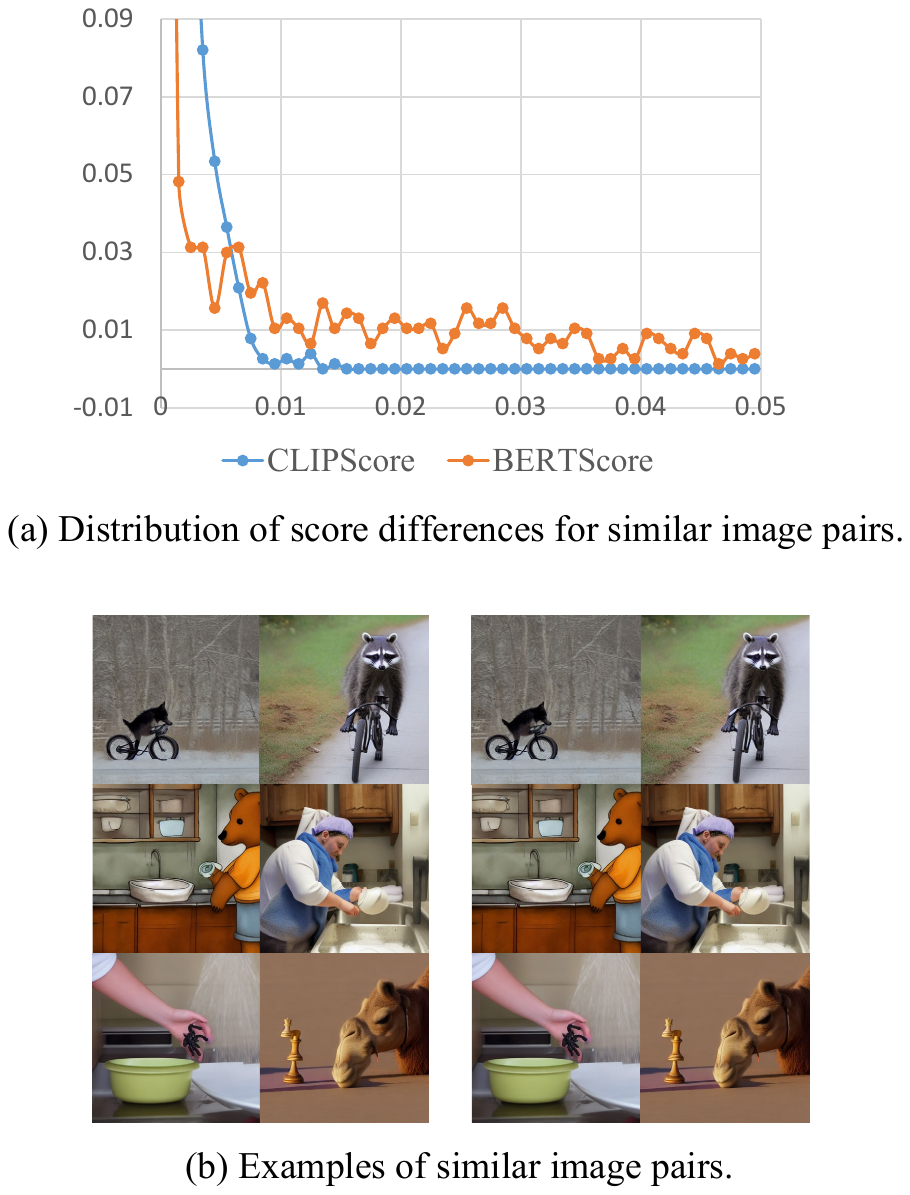}
    \caption{(a) Distribution curve of score differences for similar image pairs when evaluated by CLIPScore and BERTScore. (b) Examples of similar image pairs.}
    \label{bertdrawback}
    % \vspace{-0.4em}
\end{figure}

We create a dataset containing 768 pairs of similar images generated by diffusion models with 20 denoising steps. 
The two images in the same pair share the same states in the first 19 denoising steps, and only differ in the last denosing step. 
Some examples are shown in Figure~\ref{bertdrawback}~(b), and we can't tell the difference between them with our eyes. 
But they are different images, since their file size in JPEG format are different.  
Since images in same pairs are visually indistinguishable, they should receive similar prompt-image alignment scores. 

However, BERTScores of similar image pairs differ a lot in our observation.
Figure~\ref{bertdrawback}~(a) shows the distribution curves of score differences for similar image pairs, evaluated on CLIPScore and BERTScore. 
For CLIPScore, we can observe that almost all similar images have a score difference of less than 0.01. 
But for BERTScore, in the interval where the score difference is greater than 0.01, there are still many similar image pairs. 
As we can see from Figure~\ref{ablation}~(a), after fine-tuning the model, BERTScores of the generated images increase by 0.01-0.03. 
In consideration of accurate rewarding and evaluation, it is intolerable that the score difference of similar images is greater than 0.01. 
Therefore, we recommend using CLIPScore as reward function instead of BERTScore. 

\subsection{Introduction to Inception Score}

Following previous works~\cite{barratt2018note, bie2023renaissance, ahn2019variational, zhao2020diffaugment}, we use inception score (IS) as the metric of image diversity. Inception score is primarily applied as an evaluation metric for GANs~\cite{goodfellow2020generative}. It uses a pretrained inception v3 model~\cite{szegedy2015rethinking} to predict the conditional label distribution $P(y \mid x)$. Then the inception score is calculated as detailed in Eq.~(\ref{eq:is}): 
\begin{equation}
IS = \text{exp}(\mathbb{E}_x(KL(p(y \mid x)||P(y)))),
\label{eq:is}
\end{equation}
where $KL$ is Kullback-Leibler divergence. Traditional Inception v3 is trained only on ImageNet~\cite{5206848}, while Stable Diffusion is trained on a large-scale dataset. In real implementation, in order to better measure the diversity of images, we replace it by the image encoder of CLIP for calculating IS.
A higher inception score represents better image diversity. 

\section{More Samples}
\label{sec:more_samples}

\begin{figure*}[h]
    % \vspace{-0.4em}
    \centering
    \includegraphics[width=1.0\linewidth]{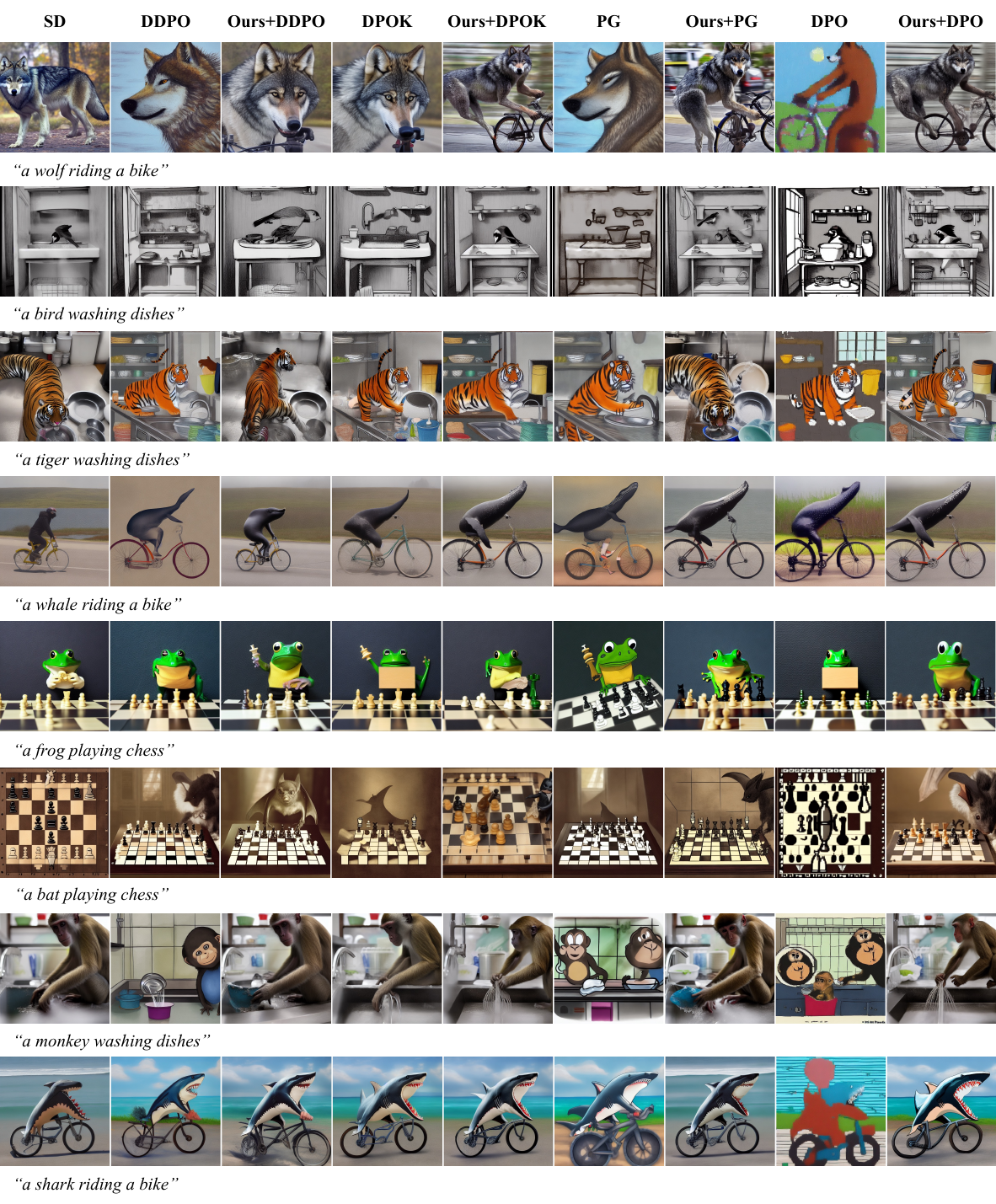}
    \caption{More samples generated by our method compared with other methods on template 1.}
    \label{fig:more_sample_t1}
    % \vspace{-0.4em}
\end{figure*}

\begin{figure*}[h]
    % \vspace{-0.4em}
    \centering
    \includegraphics[width=1.0\linewidth]{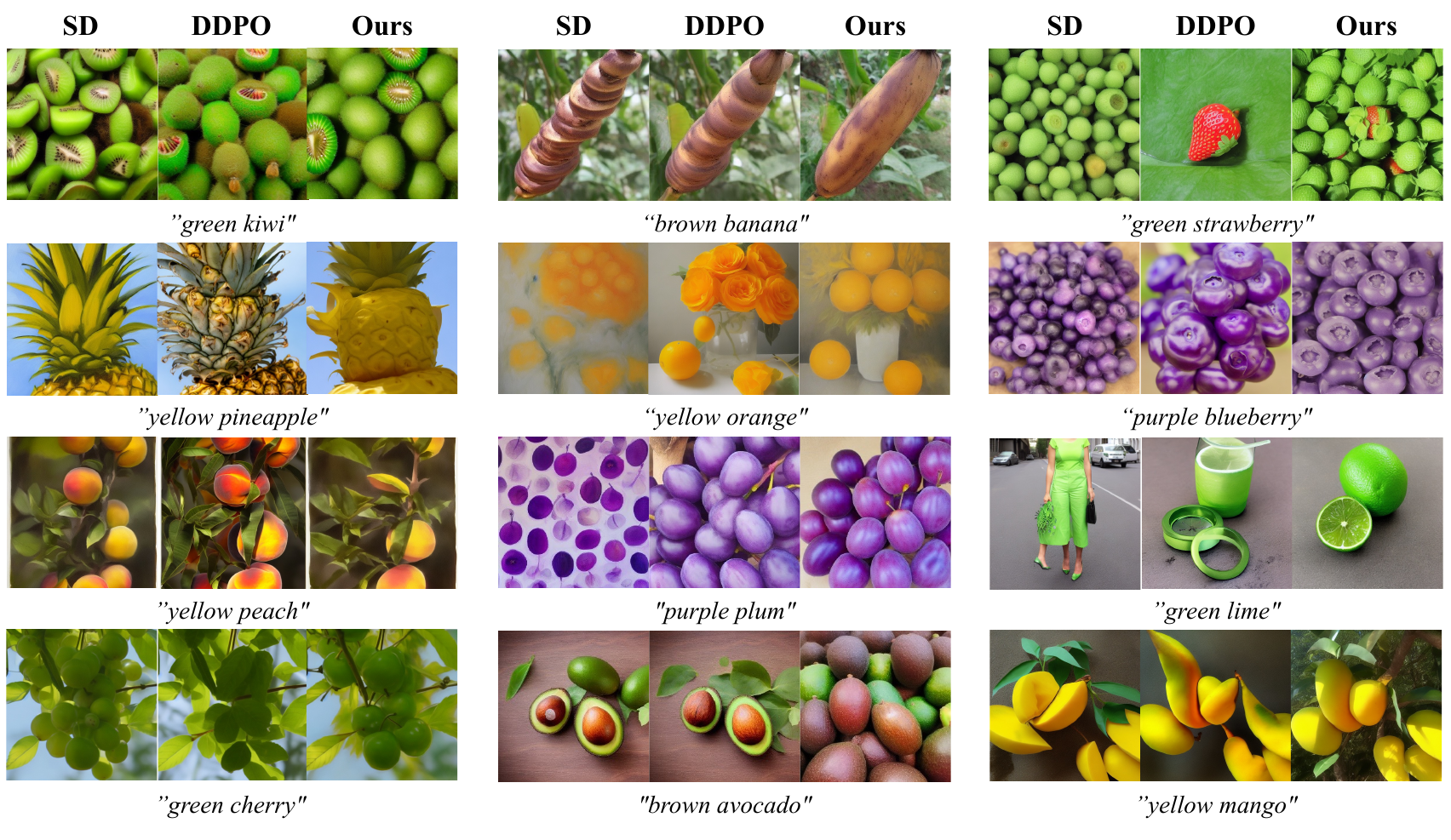}
    \caption{More samples generated by our method on template 2. }
    \label{fig:more_sample_t2}
    % \vspace{-0.4em}
\end{figure*}

\begin{figure*}[h]
    % \vspace{-0.4em}
    \centering
    \includegraphics[width=1.0\linewidth]{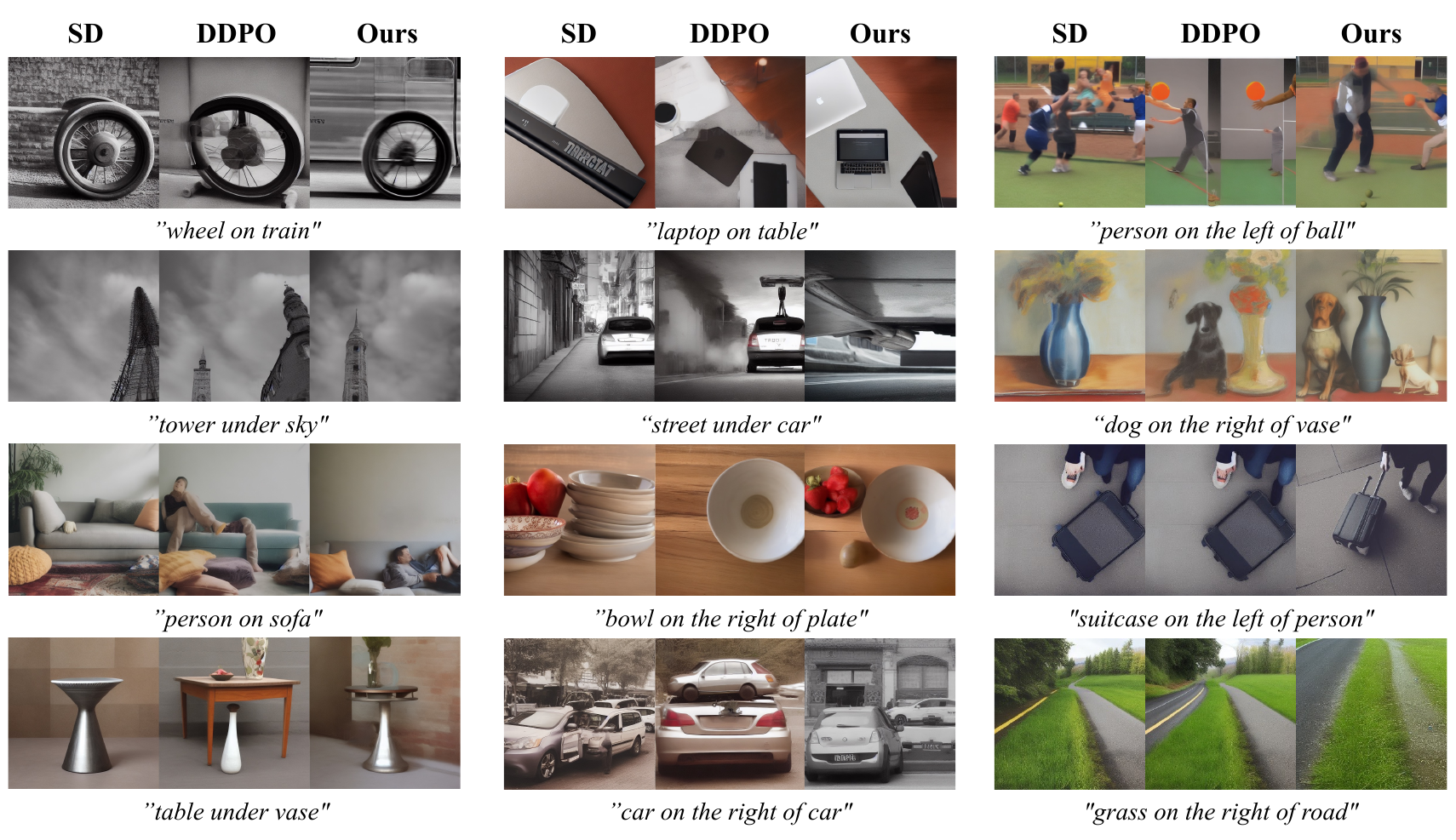}
    \caption{More samples generated by our method on template 3. }
    \label{fig:more_sample_t3}
    % \vspace{-0.4em}
\end{figure*}

\begin{figure*}[h]
    % \vspace{-0.4em}
    \centering
    \includegraphics[width=1.0\linewidth]{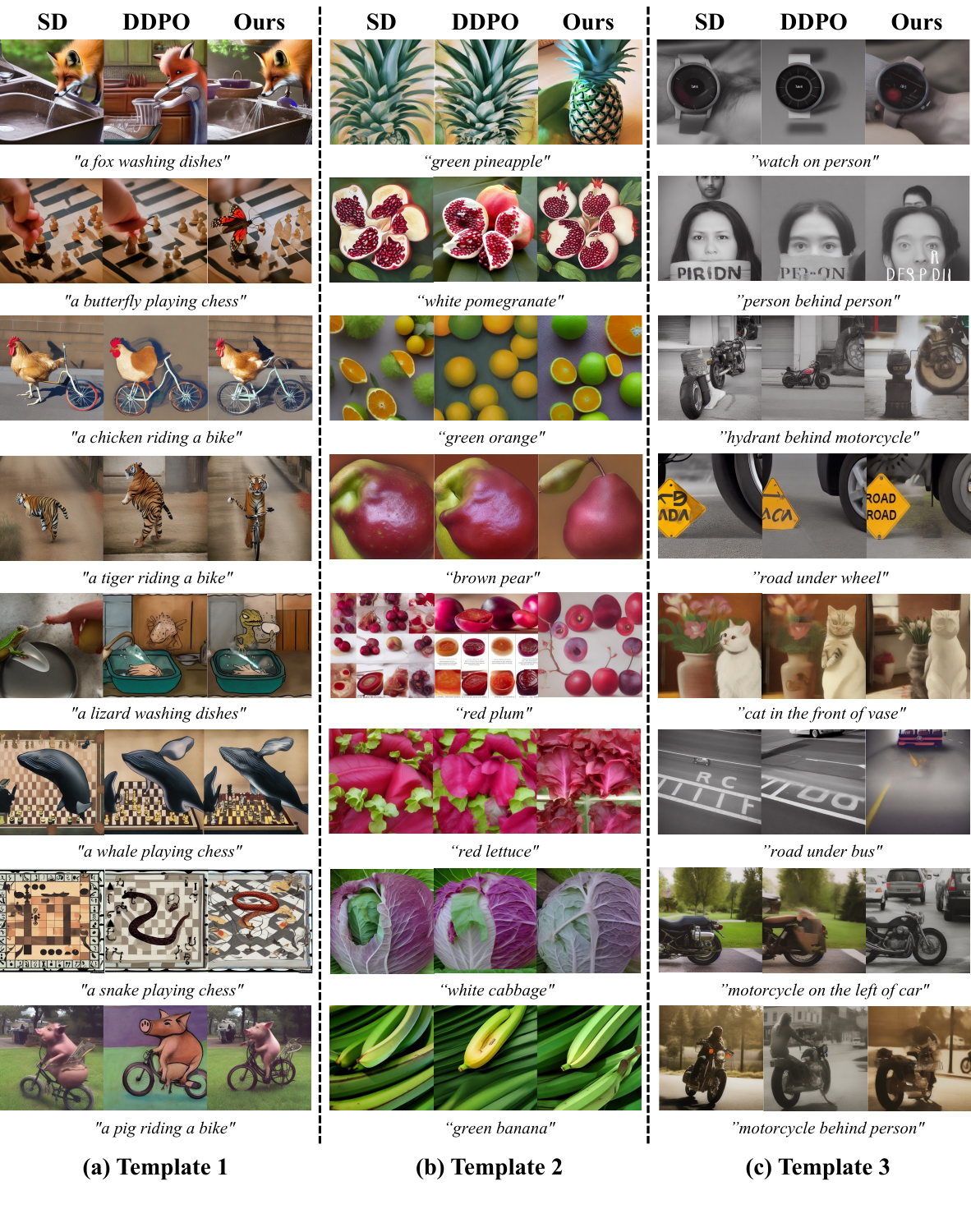}
    \caption{More samples of generalization to unseen prompts in template 1, 2 and 3.}
    \label{fig:more_sample_generalization}
    % \vspace{-0.4em}
\end{figure*}

In this section, we show more samples generated by the diffusion models fine-tuned with our method $\text{B}^2\text{-DiffuRL}$. 
Figure~\ref{fig:more_sample_t1} shows more samples generated by our method compared with DDPO, DPO, PG and DPO on templeta 1. 
Figure~\ref{fig:more_sample_t2} and~\ref{fig:more_sample_t3} show more samples from our method on template 2 and 3 respectively. 
Figure~\ref{fig:more_sample_generalization} shows more samples of generalization to unseen prompts. 

\begin{figure*}[h]
    % \vspace{-0.4em}
    \centering
    \includegraphics[width=1.0\linewidth]{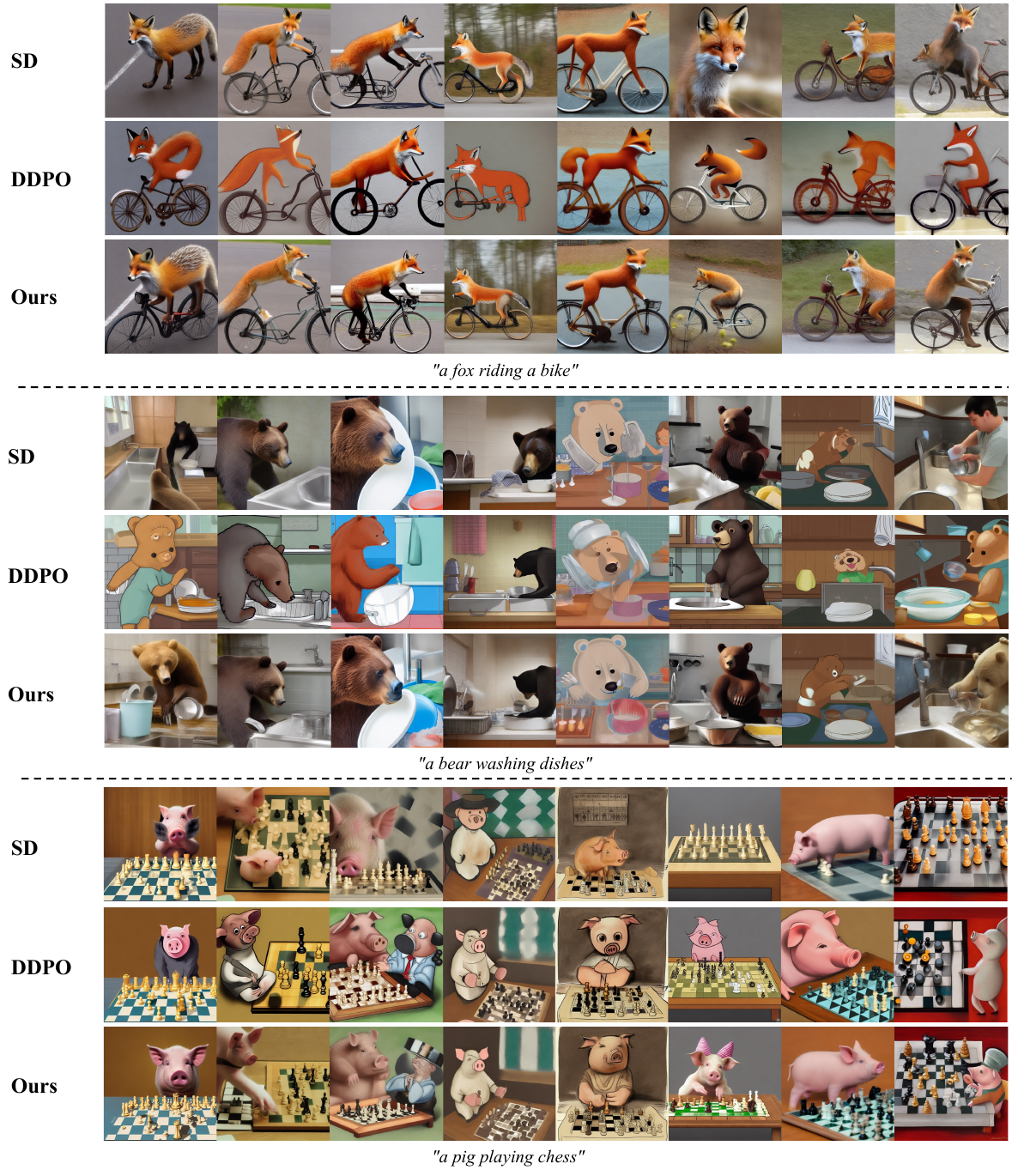}
    \caption{More samples generated by SD, DDPO and our method on three prompts. The images generated by DDPO tend to adopt a cartoon style, while those by our method tend to keep original styles of SD. These samples show that our method can help mitigating the image diversity reduction during fine-tuning.}
    \label{fig:more_sample_diversity}
    % \vspace{-0.4em}
\end{figure*}

In Figure~\ref{fig:more_sample_diversity}, more samples are generated on three given prompts to show the diversity of different methods. 
As we can see, most images generated by DDPO adopt a cartoon-like style, as described in their paper~\cite{Black2023TrainingDM}. Especially for the images generated on the prompt ``\textit{a fox riding a bike}'', almost all the background information is lost and becomes a single color. 
On the contrary, the images generated by our method can almost keep the same style as SD, mitigating the problem of diversity reduction. 

\clearpage

\section{Prompt Lists}
\label{sec:prompt_list}

In this section, we provide the prompt lists used in our experiments.
For each template, we collect one prompt list for training, and the other one for generalization test, as shown in Table~\ref{tab:template1_list},~\ref{tab:template2_list} and~\ref{tab:template3_list}. 

\begin{table*}[h]
    % \small
    \begin{tabularx}{\linewidth}{X<{\centering}|X<{\centering}|X<{\centering}}
        \toprule
        \multicolumn{3}{c}{Training list} \\
        \midrule
        a cat washing dishes & a dog washing dishes & a horse washing dishes \\
        a monkey washing dishes & a rabbit washing dishes & a zebra washing dishes \\
        a spider washing dishes & a bird washing dishes & a sheep washing dishes \\
        a deer washing dishes & a cow washing dishes & a goat washing dishes \\
        a lion washing dishes & a tiger washing dishes & a bear washing dishes \\
        a raccoon riding a bike & a fox riding a bike & a wolf riding a bike \\
        a lizard riding a bike & a beetle riding a bike & a ant riding a bike \\
        a butterfly riding a bike & a fish riding a bike & a shark riding a bike \\
        a whale riding a bike & a dolphin riding a bike & a squirrel riding a bike \\
        a mouse riding a bike & a rat riding a bike & a snake riding a bike \\
        a turtle playing chess & a frog playing chess & a chicken playing chess \\
        a duck playing chess & a goose playing chess & a bee playing chess \\
        a pig playing chess & a turkey playing chess & a fly playing chess \\
        a llama playing chess & a camel playing chess & a bat playing chess \\
        a gorilla playing chess & a hedgehog playing chess & a kangaroo playing chess \\
        \midrule
        \multicolumn{3}{c}{Test list} \\
        \midrule
        a cat riding a bike & a cat playing chess & a dog riding a bike \\
        a dog playing chess & a horse riding a bike & a horse playing chess \\
        a monkey riding a bike & a monkey playing chess & a rabbit riding a bike \\
        a rabbit playing chess & a zebra riding a bike & a zebra playing chess \\
        a spider riding a bike & a spider playing chess & a bird riding a bike \\
        a bird playing chess & a sheep riding a bike & a sheep playing chess \\
        a deer riding a bike & a deer playing chess & a cow riding a bike \\
        a cow playing chess & a goat riding a bike & a goat playing chess \\
        a lion riding a bike & a lion playing chess & a tiger riding a bike \\
        a tiger playing chess & a bear riding a bike & a bear playing chess \\
        a raccoon washing dishes & a raccoon playing chess & a fox washing dishes \\
        a fox playing chess & a wolf washing dishes & a wolf playing chess \\
        a lizard washing dishes & a lizard playing chess & a beetle washing dishes \\
        a beetle playing chess & a ant washing dishes & a ant playing chess \\
        a butterfly washing dishes & a butterfly playing chess & a fish washing dishes \\
        a fish playing chess & a shark washing dishes & a shark playing chess \\
        a whale washing dishes & a whale playing chess & a dolphin washing dishes \\
        a dolphin playing chess & a squirrel washing dishes & a squirrel playing chess \\
        a mouse washing dishes & a mouse playing chess & a rat washing dishes \\
        a rat playing chess & a snake washing dishes & a snake playing chess \\
        a turtle washing dishes & a turtle riding a bike & a frog washing dishes \\
        a frog riding a bike & a chicken washing dishes & a chicken riding a bike \\
        a duck washing dishes & a duck riding a bike & a goose washing dishes \\
        a goose riding a bike & a bee washing dishes & a bee riding a bike \\
        a pig washing dishes & a pig riding a bike & a turkey washing dishes \\
        a turkey riding a bike & a fly washing dishes & a fly riding a bike \\
        a llama washing dishes & a llama riding a bike & a camel washing dishes \\
        a camel riding a bike & a bat washing dishes & a bat riding a bike \\
        a gorilla washing dishes & a gorilla riding a bike & a hedgehog washing dishes \\
        a hedgehog riding a bike & a kangaroo washing dishes & a kangaroo riding a bike \\
        \bottomrule
        
    \end{tabularx}
    \caption{Prompt Lists for template 1. }
    \label{tab:template1_list}
\end{table*}

\begin{table*}[h]
    % \vspace{-0.1em}
    % \small
    \begin{tabularx}{\linewidth}{X<{\centering}|X<{\centering}|X<{\centering}}
        \toprule
        \multicolumn{3}{c}{Training list} \\
        \midrule
        red apple & green apple & yellow banana \\
        brown banana & orange orange & yellow orange \\
        red strawberry & green strawberry & purple grape \\
        green grape & red watermelon & green watermelon \\
        brown kiwi & green kiwi & orange mango \\
        yellow mango & green pear & yellow pear \\
        yellow pineapple & brown pineapple & orange peach \\
        yellow peach & purple plum & green plum \\
        blue blueberry & purple blueberry & red raspberry \\
        green raspberry & yellow lemon & green lemon \\
        green lime & yellow lime & green avocado \\
        brown avocado & red cherry & green cherry \\
        red pomegranate & pink pomegranate & pink grapefruit \\
        red grapefruit &  &  \\
        \midrule
        \multicolumn{3}{c}{Test list} \\
        \midrule
        yellow apple & green banana & green orange \\
        white strawberry & black grape & white watermelon \\
        white kiwi & green mango & brown pear \\
        green pineapple & red peach & red plum \\
        black blueberry & black raspberry & white lemon \\
        white lime & yellow avocado & black cherry \\
        white pomegranate & yellow grapefruit & white carrot \\
        white broccoli & yellow tomato & white cucumber \\
        brown spinach & red lettuce & yellow bell pepper \\
        white zucchini & white sweet potato & green onion \\
        green garlic & white celery & white cabbage \\
        purple cauliflower & green eggplant & purple asparagus \\
        white peas & green corn & purple green beans \\
        white brussels sprouts &  &  \\
        \bottomrule
        
    \end{tabularx}
    \caption{Prompt lists for template 2. }
    \label{tab:template2_list}
\end{table*}

\begin{table*}[h]
    % \vspace{-0.1em}
    % \vspace{-0.1em}
    % \small
    \begin{tabularx}{\linewidth}{X<{\centering}|X<{\centering}|X<{\centering}}
        \toprule
        \multicolumn{3}{c}{Training list} \\
        \midrule
        chair under umbrella & table under umbrella & car on street \\
        wheel on train & airplane on street & bag on street \\
        tree under sky & building under sky & street under sky \\
        dog on boat & tower under sky & cup on shirt \\
        person on street & laptop on table & table under laptop \\
        person on sofa & glasses on face & sofa under person \\
        table under vase & street under car & dog on the right of vase \\
        building on the right of building & suitcase on the left of person & dog on the left of person \\
        kite on the right of kite & person on the left of ball & ball on the right of person \\
        road on the left of grass & grass on the right of road & person on the left of pillow \\
        bowl on the right of plate & building on the right of truck & person on the left of bottle \\
        bottle on the right of person & box on the left of post & building on the left of building \\
        car on the right of car & truck on the right of car & car on the left of car \\
        person on the left of person &  &  \\
        \midrule
        \multicolumn{3}{c}{Test list} \\
        \midrule
        vase on table & shirt on person & watch on person \\
        jacket on person & motorcycle on road & motorcycle behind person \\
        person behind person & building behind trees & hydrant behind motorcycle \\
        trees behind grass & wheel in the front of wheel & tower in the front of train \\
        truck in the front of building & cat in the front of vase & trash can in the front of cabinet \\
        road under bus & road under building & road under wheel \\
        table under plate & person under umbrella & cone on the right of cone \\
        car on the right of umbrella & phone on the right of monitor & person on the right of bear \\
        bear on the right of person & bear on the left of person & car on the left of bus \\
        motorcycle on the left of bus & motorcycle on the left of car & road on the left of tree \\
        \bottomrule
        
    \end{tabularx}
    \caption{Prompt lists for template 3. }
    \label{tab:template3_list}
\end{table*}